# The role of neuromorphic principles in the future of biomedicine and healthcare

Grace M. Hwang[1], Jessica D. Falcone[2], Joseph D. Monaco[1], Courtney R. Pinard[3], Jessica A. Mollick[4], Roger L. Miller[5], Stephanie L. Gage[6], Andrey V. Kanaev[6], Margaret Kim[7], R. Ale Lukaszew[7], Steven M. Zehnder[7], David Rampulla[2]

[1] National Institute of Neurological Disorders and Stroke (NINDS), National Institutes of Health (NIH), The *Brain Research Through Advancing Innovative Neurotechnologies*® (BRAIN) Initiative, Bethesda, MD 20892, USA
[2] National Institute of Biomedical Imaging and Bioengineering (NIBIB), NIH, Bethesda, MD 20892, USA
[3] National Institute of Mental Health (NIMH), NIH, Bethesda, MD 20892, USA
[4] National Institute of Drug Abuse (NIDA), NIH, Bethesda, MD 20892, USA
[5] National Institute on Deafness and Other Communication Disorders (NIDCD), NIH, Bethesda, MD 20892, USA
[6] United States National Science Foundation (NSF), Directorate for Computer and Information Science and Engineering (CISE), Alexandria, VA, 22314, USA
[7] United States National Science Foundation (NSF), Directorate for Engineering (ENG), Alexandria, VA, 22314, USA

## Abstract

Neuromorphic engineering has matured over the past four decades and is currently experiencing explosive growth with the potential to transform biomedical engineering and neurotechnologies. Participants at the Neuromorphic Principles in Biomedicine and Healthcare (NPBH) Workshop (October 2024)—representing a broad cross-section of the community, including early-career and established scholars, engineers, scientists, clinicians, industry, and funders—convened to discuss the state of the field, current and future challenges, and strategies for advancing neuromorphic research and development for biomedical applications. Publicly approved recordings with transcripts (https://2024.neuro-med.org/program/session-video-and-transcripts) and slides (https://2024.neuro-med.org/program/session-slides) can be found at the workshop website.

## Introduction

Neuromorphic engineering was originally described in the 1980s as simple analog circuits that emulated the temporal processing of signals in the brain (e.g., silicon retina (1–5)). The field has since evolved from strictly small-scale hardware-specific implementations to encompass large-scale hardware (6,7), hardware/software co-design (8), AI algorithms (9), computational approaches (10), design principles (11), and bioelectronics (12,13) that collectively mimic brain and nervous system functions. A subset of neuromorphic engineering is showing promise in medical applications (14–16), including in learning circuits (17), seizure prediction (18), seizure detection (19–22), neuroprostheses (23,24), neuromodulation (25), brain-machine interface (26,27) with on-chip spike sorting (28,29), and stretchable neuromorphic wearable devices (13,30,31) (Figure 1).

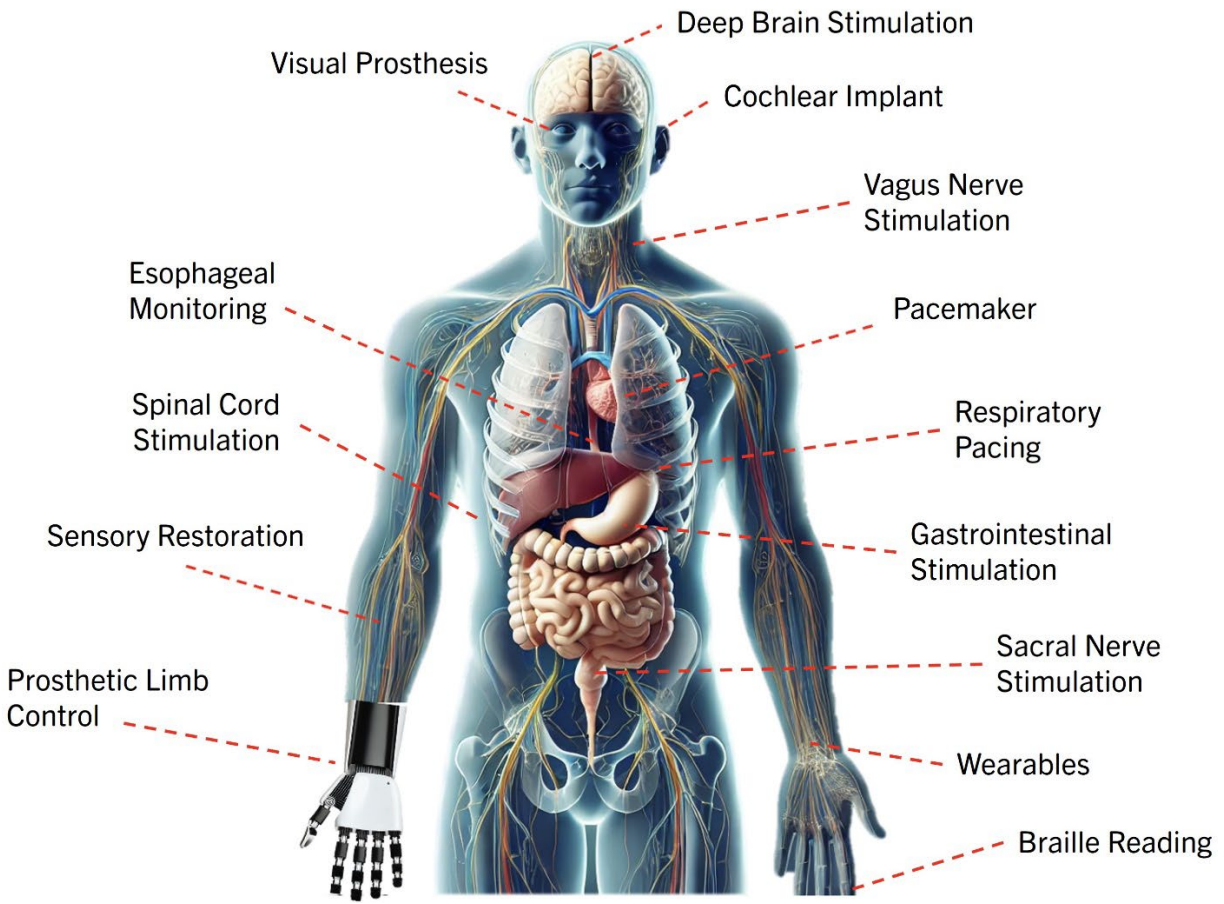

**Figure 1.** Visual overview of human body systems where future devices based on neuromorphic principles could be applied. These applications include, but are not limited to, deep brain stimulation, visual prosthesis, cochlear implants, vagus nerve stimulation, esophageal monitoring, pacemakers, spinal cord stimulation, respiratory pacing, sensory restoration, gastrointestinal stimulation, sacral nerve stimulation, prosthetic limb control, wearables, and Braille reading. Figure credit: adapted with permission from Elisa Donati's Neuromorphic Principles in Biomedicine and Healthcare workshop presentation in Session 2.

Inspired by recent advances in neuromorphic engineering for health, National Institutes of Health (NIH) program officials—representing the NIH *Brain Research Through Advancing Innovative Neurotechnologies*® (BRAIN) Initiative, the National Institute of Biomedical Imaging and Bioengineering (NIBIB), and the National Institute of Neurological Disorders and Stroke (NINDS)—collaborated with U.S. National Science Foundation (NSF) program directors—representing the Directorate for Engineering (ENG) and the Directorate for Computer and Information Sciences (CISE)—to organize the Neuromorphic Principles in Biomedicine and Health (NPBH) hybrid workshop on 21-22 October, 2024, in Baltimore, Maryland, USA. See Table 1 for a list of the invited speakers, co-chairs, panelists, and guiding questions. The NPBH workshop had 171 participants (84 in-person; 87 virtual), ranging from early-career to established scholars and including biomedical engineers, neuromorphic engineers, neuroscientists, material scientists, clinicians, small businesses, federal funders, and private investors. The participants convened with the goal to explore the intersection of neuromorphic engineering with healthcare applications. A forward-looking community roadmap will be published separately by workshop attendees. This article serves to summarize workshop presentations and discussions and highlight opportunities for neuromorphic engineering in neurotechnology and biomedical engineering. Section 1 (Topics Covered and Major Discussions) and Section 2 (Open Questions and Critical Challenges (Barriers) are based solely on workshop discussions. Section 3 (Outlook and Vision for Next Steps) comprises perspectives from workshop participants and program representatives that were synthesized following the workshop.

**Table 1**: Speakers, Co-chairs, Panelists, and Guiding Questions

| Speaker Name | Title of Talk | Co-chairs/ Panelists | Guiding Questions |
|---|---|---|---|
| **Day 1 – 21 October, 2024** | | | |
| Timothy Denison (University of Oxford, England) | Keynote: Towards "Neuro/Physio-morphic" Devices [Generation 3?] | | |
| *Session 1 Topic 1 - Epilepsy & Cortical Disorders (Neuromodulation)* | | | |
| William Stacey (University of Michigan) | Data Driven classification without a gold standard: epileptic HFOs | Session 1 Co-chairs: Grace Hwang (NIH/NINDS), Jessica Falcone (NIH/NIBIB) | 1. If you had a low power, implantable chip that could run any algorithm (real-time, closed-loop, machine learning (ML), artificial intelligence (AI), etc.), how would you use it? What are the encoding and decoding challenges in epilepsy, movement, and psychiatric disorders? |
| Sridevi Sarma (Johns Hopkins University (JHU)) | A Resting State EEG Biomarker Points to Seizure Onset Zone | Panelists: Sydney Cash (Harvard Massachusetts General Hospital), Nathan Crone (JHU), Kendall Lee (Mayo Clinic), Giacomo Valle (Chalmers University of Technology, Sweden), Chris Rozell (Georgia Institute of Technology, GIT) | 2. What strategies (materials, technology, algorithms, theory) can we use to interface with the brain in a more neuromorphic manner? Is it critical for neuromodulation (hardware or software) to mimic the brain? Why? |
| Viktor Jirsa (Aix-Marseille Université, Inserm, France) | Virtual Brain Twins for Medicine - Encephalomorphic Systems | | 3. What impact could major neuromodulation companies have in this process? What is needed for clinical translation? |
| *Session 2 Topic 2 - Human-machine interfaces (Prosthetics)* | | | |
| James Cotton (Northwestern University) | Clinical Gaps and Understanding Real World Outcomes for Prosthesis Users with AI Technologies | Session 2 Co-chairs: Luke Osborn (Case Western Reserve University), Francisco Valero-Cuevas (University of Southern California) | 1. How can neuromorphic principles impact the ability to convey complex or high-resolution sensory information through neural interfaces? What's the current limitation? Processing power, sensors, neural interfaces, physiological understanding? |
| Ranu Jung (University of Arkansas) | Convergence: Deploying Neuroelectronic Interfaces | Panelists: Xing Chen (University of Pittsburgh), Nitish Thakor (JHU), Helen Huang (North Carolina State University), Vikash Gilja (University of San Diego | 2. In what ways could a low power system, which adapts to and evolves with the user, benefit human- |
| Elisa Donati (University of Zurich & ETH | Neuromorphic Hardware for Closed-loop in Healthcare | | |

| Speaker Name | Title of Talk | Co-chairs/ Panelists | Guiding Questions |
|---|---|---|---|
| Zurich, Switzerland) | | (UCSD)), Shih-Chii Liu (University of Zurich & ETH Zurich, Switzerland) | machine interfaces or rehabilitation tools? <br> 3. What are the benefits for using neuromorphic computing in neural prosthesis? Does it bring extra benefit for robust daily use? Are there unknown functional benefits? <br> 4. Is there an opportunity to use neuromorphic principles to reveal mechanisms that give rise to function, and explain dysfunction, while pointing a path to clinical treatment? |

**Day 2 – 22 October, 2024**

| | | | |
|---|---|---|---|
| Zhenan Bao (Stanford University) | Keynote: Learning from skin: from Materials, Sensing Functions to Neuromorphic Engineering | | |

*Session 3 Topic 3 - Materials for Neuromorphics (Devices)*

| | | | |
|---|---|---|---|
| Duygu Kuzum (UCSD) | Innovating Beyond Electrophysiology: Multimodal Neural Interfaces and Neuromorphic Co-Design | Session 3 co-Chairs: Duygu Kuzum (UCSD), Shantanu Chakrabarty (Washington University St. Louis) Panelists: Gina Adam (George Washington University), Jennifer Hasler (GIT), Gert Cauwenberghs (UCSD), Ulkuhan Guler (Worcester Polytechnic Institute) | 1. What are the opportunities in using new materials and device technologies? How will these interface with the tissue (biocompatibility) and complementary metal-oxide semiconductor (CMOS) electronics? <br> 2. Is neuromorphic the only solution to power, bandwidth and throughput problems for implants? <br> 3. What are the challenges in manufacturing neuromorphic chips for biomedicine? Cost? <br> 4. What are the benefits of on-chip learning or adaptation for neuromorphic hardware? <br> 5. What are the challenges in translation of neuromorphic to the clinic? What role should industry play in that? |
| Tony Lewis (BrainChip) | Flow Machines and Neuromorphic Devices | | |
| Deji Akinwande (University of Texas Austin) | Towards Health Monitoring & Computing using Advanced Materials | | |

*Session 4 Topic 4 - Medical Imaging, Wearables and Analysis*

| | | | |
|---|---|---|---|
| Sihong Wang (University of Chicago) | Bio-electronic interfaces: from high fidelity biosensing to integrated neuromorphic edge computing | Session 4 co-Chairs: Ralph Etienne-Cummings (JHU), Jennifer Blain Christen (Arizona State University) Panelists: Andreas Andreou (JHU), Robert Stevens (JHU), Pamela Abshire (University of Maryland, College Park), | 1. If you had a low power, implantable chip that could run any algorithm (real-time, closed-loop, ML, AI, etc.), how would you use it? <br> 2. What new strategies (materials, technology, algorithms, theory) can we utilize? <br> 3. How can neuromorphic principles make these ideas a reality? |
| Erika Schmitt (Cambrya) | Novel neuromorphic algorithms for adaptive learning at the edge | | |

| Speaker Name | Title of Talk | Co-chairs/Panelists | Guiding Questions |
|---|---|---|---|
| J. Brad Aimone (Sandia National Laboratories) | From new neurons to new chips: a path to a new way of thinking about the brain? | Amparo Güemes (University of Cambridge, England) | 4. What impact/role does major implantable / wearables companies have in this process? What is the pathway for clinical translation? |

Publicly approved recordings with transcripts can be found on the NPBH workshop website at https://2024.neuro-med.org/program/session-video-and-transcripts. Presentation slides can be found at https://2024.neuro-med.org/program/session-slides. Appendix 1 contains a synopsis of the meeting. Appendix 2 presents the panelists' Position Statement on the Needs, Challenge, Contribution, Impact and Investment for the field.

The topics discussed at this workshop have reverberated across related meetings sponsored by the NIH and the NSF including: (1) the 23 October 2024 NIH Wednesday Afternoon Lecture Series (WALS) lecture (https://videocast.nih.gov/watch=55007) in which Brad Aimone introduced his Sandia research group's preliminary implementation of the adult *Drosophila melanogaster* connectome on the Loihi neuromorphic platform comprising 140K neurons and 50M synapses (32); (2) the 12-13 November 2024 NIH BRAIN NeuroAI Workshop (33,34) (Day 1 videocast, https://videocast.nih.gov/watch=55160; Day 2 videocast, https://videocast.nih.gov/watch=55262); (3) the 27-29 August 2025 NSF-sponsored NeuroAI+Beyond Workshop (NSF Award No. 2533276; (35)); (4) the 27 August 2025 NIH BRAIN Multi-Council Working Group meeting (https://videocast.nih.gov/watch=56872) at which the NeuroAI Innovation Domain was publicly introduced by NIH BRAIN Director John Ngai as part of the BRAIN Initiative's anticipated future research roadmap (36); (5) the 13 November 2025 NIH/BRAIN Minisymposium on Neuromorphic Engineering for Clinical Care held at the BRAIN-sponsored 12[th] Annual Institute of Electrical and Electronics Engineers (IEEE) Engineering in Medicine and Biology Society's (EMBS) International Conference on Neural Engineering and Rehabilitation (NER25) (https://neuro.embs.org/2025/nih-brain-initiative-minisymposium/; https://www.youtube.com/watch?v=ljsdScNvrKg).

## Topics Covered and Major Discussions

The workshop emphasized bidirectional knowledge exchange between engineers and clinicians, with particular focus on whether and how neuromorphic approaches can address critical challenges for clinical applications. Throughout the meeting, participants highlighted the importance of focusing on real-world impact rather than merely mimicking biological structures and dynamics. This perspective was reinforced through case studies of devices that have successfully translated to clinical use, such as adaptive deep brain stimulation (DBS) for Parkinson's disease (PD) and sensory feedback systems for prosthetic limbs.

# Current Understanding and State-of-the-art

## Day 1 Keynote: Towards "Neuro/Physio-morphic" Devices [Generation 3?]

Timothy Denison presented on the opportunity and challenges for using electronic hardware and software to intelligently stimulate the nervous system in real time to ameliorate symptoms. He focused on adaptive DBS systems like the Medtronic stimulator (e.g., Percept$^{TM}$ (37,38), Summit RC+S (39)) that have shown clinical benefits for PD patients. He emphasized the importance of not just neuromorphic but also closed-loop "physiomorphic" approaches that use physiological markers to adjust stimulation in real time to better match individual brain states. These systems utilize dual-threshold homeostatic approaches inspired by natural physiological systems, which outperform simpler threshold-based controls. Algorithmic processing is performed on external systems where parameters can be updated through telemetry. Although these systems do not currently employ neuromorphic hardware or algorithms, Denison explained that these physiomorphic approaches are inspired by human physiology. He presented a vision for "Generation 3" devices that would model and restore physiology, noting that current neuromodulation devices lack predictive capabilities. Denison further noted that algorithms are often rudimentary, due to limited power budgets on current devices. While neuromorphic technologies may confer advantages in power efficiency, they need to demonstrate advantages over classic digital signal processing and patient-relevant performance metrics. A great way to engage the community on novel algorithm design could be through "battle of the algorithms" competitions that evaluate different devices running different algorithms for different diseases, allowing researchers to monetize their ideas.

## Topic 1 - Epilepsy & Cortical Disorders (Neuromodulation)

William Stacey explained that current surgical decisions for intractable epilepsy largely rely on empirical observations and visual analysis of electroencephalography (EEG) patterns derived from decades-old methodologies. Existing neuromodulation devices, such as DBS, responsive neurostimulation (RNS), and vagus nerve stimulation (VNS), were not optimized for epilepsy treatment and stimulation parameters were initially borrowed from other applications. Some closed-loop RNS devices sense and stimulate with a four-year battery life but suffer from extremely low processing power. For example, only five minutes of continuous data from a few channels are typically saved over the course of 24 hours (25). Identifying biomarkers to determine epileptogenic zones remains an active area of research owing in part to the lack of gold standard validation.

Three promising methods for treating epilepsy were presented, including: (1) high-frequency oscillations (HFOs) (by Stacey) from intracranial EEG (iEEG) (40–42); (2) resting state iEEG (by Sridevi Sarma) analysis devoid of seizure data (43); (3) virtual brain models (by Viktor Jirsa) constructed from personalized neuroimaging datasets (44–47). Stacey explained that neuromorphic technologies and approaches are rarely used in epilepsy in the United States. Demonstrated uses include real-time seizure prediction (IBM TrueNorth (48)) and real-time HFO detection from iEEG in Zurich Switzerland (22,49).

## Topic 2 - Human-machine interfaces (Prosthetics)

Three talks were presented on clinical perspectives of neural interfaces and rehabilitation. James Cotton started the session with emphasis on patient perspectives, revealing that patients prioritize comfort, reduced pain, and psychological factors over advanced technical features, with studies showing socket fit issues and skin problems as patient-reported priorities (50).

As described by Ranu Jung (51), the fundamental challenge of biohybrid systems remains largely unsolved: biological systems are non-linear, non-stationary, multi-scale and plastic, yet current technologies remain relatively fixed. Research efforts to address this problem in prosthetics date back to 2006 and were originally funded as part of an NIH Bioengineering Research Partnership Program, DARPA's RE-NET and HAPTIX programs, and the Joint Warfighter Medical Research Program. For example, a fully implanted system was developed that provided sensory feedback from sensors in a prosthetic hand via wireless communication to implanted electrodes to directly stimulate nerves (52,53). In one case, a prosthetic hand participant was able to feel sensation in his missing hand for the first time in 74 years (54). The prosthetic technologies presented by Jung were not neuromorphic, but served to represent an assessment of the state of the art.

Elisa Donati presented neuromorphic hardware approaches that specifically enabled efficient sensory feedback by directly stimulating nerves with patterns that mimic natural sensation (24). These systems have comparable accuracies to deep learning-based approaches, yet use significantly fewer parameters and less power (55,56). Donati also described a "Neuromorphic Twin" as an emerging tool that physically emulates the somatosensory cortex of human patients to create neurostimulation patterns. This emulator could improve the quality of restored feedback for neural prostheses by better reproducing natural sensations (57). Other promising biomedical neuromorphic interfaces covered in this session include cardiac activity monitoring (58) (from Donati) and real-time personalized respiratory pacing (59) (from Jung).

## Day 2 Keynote: Learning from skin: from Materials, Sensing Functions to Neuromorphic Engineering

Zhenan Bao provided an update on a collection of skin-like electronic polymers and materials with enhanced electrical capabilities compared to traditional semiconductor materials. She presented front-end processing pressure sensors on a stretchable substrate with over 1,000 transistors in the integrated circuit. These pressure sensors exceeded human tactile resolution (better than 500 microns), which was demonstrated with improvement in Braille reading (60). Another technology, termed "neuromorphic skin", had no rigid electronic components and emulated the sensory feedback functions of biological skin, including multimodal reception, nerve-like pulse-train signal conditioning, and closed-loop actuation (13). Bao noted that prototyping these nonconventional materials presents many challenges, one of which is testing long-term biocompatibility. Bao concluded that other key challenges for the community are prototyping nonconventional materials and identifying which biomedical applications would benefit most from neuromorphic or hybrid design approaches.

## Topic 3 - Materials for Neuromorphics (Devices)

Two talks focused on novel materials while a third talk was about a commercial neuromorphic processor. For neural interfaces, Duygu Kuzum described how transparent graphene-based microelectrodes enabled multimodal experiments by simultaneously collecting electrophysiological recordings while allowing optical imaging and stimulation without crosstalk (61). As an example, she described resistive RAM-based neuromorphic processors developed in her lab that performed efficient spike-sorting with 30-50 times less energy consumption than traditional approaches (62). Kuzum showed how principles of neuromorphic co-design could be applied to decode individual cellular calcium activity in deeper brain layers using only surface potentials measured from graphene-based electrodes and neuromorphic computational models with recurrent neural networks. This technique successfully expanded the spatial reach of deployed electrodes (61).

Deji Akinwande presented two-dimensional (2D) materials (e.g., graphene, less than a nanometer thick) for use in "organ-tronics" (*in vivo* sensing) and "skin-tronics" (direct skin interfacing). Applications included electrophysiological monitoring (63), implantable sensing (pacemaker (64)), and a wrist-worn bioimpedance sensor for monitoring blood pressure with clinical-grade accuracy (65). He also presented "Atom-resistors" (memristors in atomically thin materials) that demonstrate both binary and analog resistance switching (66) in neural networks, with classification accuracy approaching theoretical limits in a Modified National Institute of Standards and Technology database (MNIST) digit recognition task.

On the commercial side, Tony Lewis described BrainChip's Akida architecture, which operates as an event-based, digital, near-memory computing system using weighted spikes. Akida achieved state-of-the-art results in a myriad of tasks (gesture recognition, eye tracking, denoising, and audio keyword spotting) while operating at extremely low power (0.25 milliwatts). Lewis presented data showing that BrainChip's Temporal Event-Based Neural Networks processing unit achieved better perplexity scores than state-of-the-art large language models (e.g., Mamba and Llama 3.2) using a fraction of the training data and half the model size.

## Topic 4 - Medical Imaging, Wearables and Analysis

One talk focused on wearables, while two talks focused on algorithms. Sihong Wang presented stretchable, soft semiconductor polymers that yielded stable signal quality, directly adhered to tissue surfaces (such as a beating heart), and suppressed the foreign body response by incorporating immunomodulatory properties into the polymers (64). His presentation also suggested the need to shift the processing of sensor data from off-chip, cloud computing to on-chip, edge computing. Wang then demonstrated a fully stretchable, wearable array for on-chip classification of electrocardiogram (ECG) signals with electrochemical transistor-based neuromorphic hardware (30,67).

Brad Aimone presented neuromorphic algorithms that leveraged structural relationships from connectomes rather than relying solely on sequential time-based analyses, and which excelled at sampling problems (68) and finite element approaches (69). His neuromorphic approach to solving partial differential equations showed dynamics remarkably similar to biological neural recordings, despite drastic differences in architectures. Aimone noted that both neuromorphic approaches and the biological brain utilized heterogeneity, parallel processing, and spatial temporal processing. In a recent preprint, Aimone additionally highlights these biological features as critical to developing theoretical frameworks to characterize the limits and capabilities of neuromorphic computing (70). On the commercial side, Erika Schmitt described Cambrya's work developing neuromorphic algorithms to solve the auditory blind source separation problem using density networks that do not require prior training (71).

## Regulators and Funders Session

Program staff from the NSF, NIH, Army, and Defense Advanced Research Projects Agency (DARPA) presented relevant funding opportunities and competitions to encourage participant submissions. Additional details can be found in Appendix 1; funders' presentations can be retrieved from https://2024.neuro-med.org/program/session-slides.

The United States Food and Drug Administration (FDA) provided an overview of programs for overseeing novel medical devices, explaining a three-tier classification system and highlighting the need to engage early through the Q-submission process. Key considerations raised during discussions throughout the meeting included: (1) difficulty of validation of adaptive systems; (2) security and safety, including the need for failsafe mechanisms in general, and cybersecurity in implantable systems; (3) manufacturing verification, including challenges in verifying neuromorphic hardware manufacturing; and (4) educating researchers about the regulatory process. It was noted that the FDA Guidance document on Physiologic Closed-Loop Control Devices (72) and the International Electrotechnical Commission's (IEC) Document # 60601-1-10 (73) were good starting points for device developers. The OpenGPU Foundation also presented their interest and rationale for investing in hardware that demonstrated advantages in temporal computing, noisy stochastic computing, energy efficiency, and multidimensional computing. OpenGPU additionally discussed its aims to create standardized interfaces between hardware and "wetware" (e.g., organoids).

## Open Questions and Critical Challenges (Barriers)

The neuromorphic biomedical field faces several significant challenges that must be addressed to realize its full potential. This section provides a thematic organization of challenges discussed and debated by NPBH workshop panelists and participants.

### Clinical Challenges

*Epilepsy, neurological, and psychiatric conditions*

Clinicians expressed major challenges inherent to the study of epilepsy and the identification of the epileptogenic zone. The state-of-the-art devices used to stimulate specific brain regions to predict or abort seizures are not sufficient. Implantable devices suffer from significant limitations in sampling rate and bandwidth, bit precision, and computational capabilities, hampering research progress. Clinicians noted that although the field is moving away from a one-setting-fits-all model and towards personalizing the settings on devices to meet the needs of individuals, there remains an urgent need for devices with more channels, customizable sampling rates, and greater signal-to-noise ratios in fully implantable wireless form-factors. Over the course of the workshop, panelists identified a critical need for intermediate biomarkers that could provide earlier indications of treatment efficacy prior to the change in clinical symptoms, particularly in psychiatric applications.

*Data collection limitations*

The limitations of current data collection were also discussed. Some devices do not record and save data continuously, and high-resolution data is primarily available only during inpatient monitoring rather than in ambulatory settings. Participants provided specific technical requirements for neural recordings in ambulatory and inpatient acute abilities, including customizable electrode placement, stimulation parameters, sampling rates (0.5–4,000 Hz), and continuous data collection. Participants identified another critical gap in the need for time-synchronized, simultaneous multimodal recording with high spatiotemporal resolution for electrical and chemical signals including potassium, oxygen, and various neurotransmitters. Developing this multimodal data collection capability would be a critical improvement for inpatient acute settings.

### Engineering Challenges

Whether neuromorphic approaches offer real-world advantages over conventional methods remains an open question. Some participants argued that neuromorphic systems excel at specific applications that recover function (e.g., limb loss, sight or hearing restoration, peripheral nerve injury, spinal cord injury) and require ultra-low power or real-time adaptation. Others suggested focusing on demonstrated capabilities rather than philosophical adherence to biological principles. While neuromorphic technologies may be advantageous in power efficiency or latency, workshop participants asserted that there must be demonstrated benefits in various design considerations, including manufacturability, reliability testing, management of signal drift in analog systems, system verification, risk management, and optimization across multiple parameters (e.g., not only power consumption).

The dichotomy of off-chip vs. on-chip computing was raised multiple times. Participants discussed whether future neuromorphic circuits should communicate wirelessly with sensors and stimulators or be implanted together as an integrated system. The latter approach was preferred according to multiple workshop participants who mentioned that radio telemetry remains expensive compared to on-chip computation, with telemetry consuming upwards of 80% of the power budget (74,75). The development of autonomous (telemetry-free) closed-loop neuromodulation systems remains a critical challenge that neuromorphic technologies are poised to tackle (25).

For medical devices, some participants expressed cybersecurity concerns for systems that require continuous communication with the cloud. Neuromorphic approaches were suggested to be inherently less hackable due to the potential for self-sufficient, autonomous operation and on-board processing.

For brain-machine interfaces (BMIs), several participants asserted that key technical challenges include adapting to signal variability in neural interfaces. Recordings can change dramatically within weeks due to shifts in electrode position or cellular responsivity, which could be addressed through on-chip learning and adaptation. Participants noted that the concept of "degeneracy"— that biological systems like the human brain achieve resilience by adaptively selecting from multiple distinct pathways and structures to implement any given function or behavior — presents a fundamental complication for the encoding and decoding of neural signals.

Participants outlined other technical challenges that remain for neural interfaces, including tissue damage during electrode implantation, limited spatial resolution of recordings, and bulky head-stage equipment for optical imaging.

Regarding hybrid neural interfaces, one participant noted that the field currently lacks a cross-mapping for understanding bidirectional control between biological and artificial silicon neurons. The participant suggested that a key measure of the functional embodiment of an artificial neuron is whether biological neurons integrate with it and treat it "as one of their own". For example, an artificial embodiment of a given neuronal cell type could be engineered to enhance repair functions at the neural interface which improve measures of bidirectional control.

## Materials Challenges

Many participants expressed the view that integrating neuromorphic devices at the biological interface remains a challenge. For both wearable and implanted devices, seamless interfaces require biocompatible, stretchable, conformable, flexible, and often biodegradable materials, which elicit a minimal immune response. Participants viewed the organ, location, and duration of the interface to be key factors to consider when evaluating a device's biocompatibility and stability. There was consensus that prototyping and assessing the long-term biological success of novel, nonconventional materials would be difficult.

## Benchmarks vs. Datasets

Workshop participants identified the need to develop specific, standardized datasets rather than focusing on benchmarks, which some participants found to be restrictive (refer to (76) for an example). Aimone argued that benchmarking, despite its popularity in ML/AI development, would be counterproductive for the neuromorphic community as developers would be forced to use outdated paradigms while the field of neuroscience moves forward. Andreou offered historical context from speech processing, noting that benchmarks initially helped standardize evaluation. However, innovation was eventually constrained since researchers optimized specifically for benchmark datasets rather than solving real-world problems. The discussion highlighted the precarious balance between finding value in standardized datasets while avoiding restrictive benchmarks.

Workshop participants recognized the need for datasets to characterize neuroscience principles and enable the co-design of materials, devices, and algorithms that compute locally. This critical challenge will require closed-loop integration between sensing, computing, and actuation. Another critical challenge that arose in discussions was the availability of time-synchronized multi-modal datasets for a single task. Finally, the development and use of standardized metrics, including energy efficiency, compactness, noise, and robustness of the hardware and software, was also highlighted as a challenge.

## Infrastructure Challenges

*Access to fabrication facilities, design tools, standards*

Infrastructure limitations constrain progress in the field, with many workshop participants calling for better access to fabrication facilities, design tools for asynchronous circuits, standardized interfaces between hardware and "wetware" (biological systems), and better sharing of protocols especially for novel materials. There was discussion about limited access to fabrication facilities and a strong call to establish programs similar to the historical MOSIS (77) initiative that provided free chip manufacturing for academic researchers. (The authors note that "MOSIS 2.0" was re-launched in 2023 as a component of a California innovation hub supported by the U.S. CHIPS and Science Act of 2022; see https://www.mosis2.com/about-us for details.) Participants suggested expanding beyond silicon chips to incorporate new materials and structural complexity (e.g., bendable bioelectronics) with a supporting infrastructure equivalent to MOSIS for these novel approaches. Participants expressed the view that development would accelerate if tools for designing asynchronous circuits and transferring spiking neural networks between hardware platforms were available. Access to standardized testing methods for novel components like memristor arrays was noted by participants as a limitation. Protocol sharing to eliminate the need to start from scratch, especially when developing novel biohybrid neuromorphic materials was highlighted multiple times by participants. The gap between the design tool ecosystems available for traditional digital signal processing versus neuromorphic computing was also emphasized as a barrier to broader adoption.

## Manufacturing Challenges

Manufacturing challenges were discussed by participants, including split foundry issues (e.g., where the lower layers and upper layers are manufactured in different locations) and tool-chain limitations. The high cost of chip fabrication was also raised, with participants' estimates ranging from $2–30 million to produce a new chip design. Some participants argued that a few thousand dollars on a multi-project

wafer (MPW) run could yield sufficient production for healthcare applications. For example, after the workshop, Andreou clarified that minimum orders for MPW through Europractice or Muse were estimated at $5,000 or $6,250 depending on process node and that Europractice also supports fabrication of circuits using organic semiconductors and flexible substrates. Different participants noted on multiple occasions that neuromorphic applications can be developed with standard CMOS technology, which is much more widely available. Another challenge raised by participants was developing testing methods for novel components, like memristor arrays. Participants further noted that it was difficult to predict chronic reliability in harsh biological environments, where performance can be impacted by temperature fluctuations, foreign body response, and material degradation.

## Commercialization

*Regulatory Challenges*

Participants noted that the regulatory pathway presented additional complexity, particularly for adaptive systems that learn and change over time. This raised questions about how to quantify the benefits of adaptation and establish appropriate regulatory frameworks that enable responsible innovation. The importance of fail-safe mechanisms in adaptive systems to ensure devices revert to baseline functionality if problems arise was raised multiple times by participants. The similarity to autonomous vehicles with different levels of autonomy was also made. One participant noted that similar concerns arose when NeuroPace RNS was first launched; dose limits were employed as guardrails to limit the risk to patients should the device go awry. Many participants expressed the need for better education for researchers about regulatory process.

*Clinical Translation Challenges*

Participants expressed concerns that clinical translation is hindered by long investment horizons that do not align with typical business expectations. There are difficulties in demonstrating improved real-world function, beyond laboratory performance. Participants discussed a disconnect between what researchers prioritize (e.g., advances in technical features) versus what patients value (e.g., improved comfort, reduced pain, and avoiding embarrassment from device failures). Patient privacy and data security considerations were also mentioned as challenges for clinical translation.

*Investment Challenges*

Participant consensus pointed to a "chicken and egg" problem in investment — developers need data to justify investment in novel devices, but gathering this data can require substantial initial investment in the devices. Some participants stated that this has led to a conservative technology development approach in which companies incrementally improve their established technologies rather than pursuing innovative designs. Multiple industry participants noted that long investment horizons do not align with typical business expectations.

## Transdisciplinary Communication and Education

One of the most persistent themes throughout NPBH workshop discussions was the need for transdisciplinary education to bridge the communication gap between engineers, scientists, and clinicians. For example, Gina Adam stressed the importance of interdisciplinary collaboration and education, while Nitish Thakor noted that established neuroscience communities — often rooted in experimentally-driven, mechanistic, and reductionist approaches — may find neuromorphic concepts unfamiliar or difficult to contextualize within their frameworks. Participants also called for developing transdisciplinary teams that include neuromorphic engineers, material scientists, neuroscientists, clinicians, data scientists, and regulatory experts. Amparo Güemes raised the question of whether one lab should house many domains of expertise or specialized labs should be encouraged to collaborate, a decision which is particularly impactful for hiring when setting up a lab.

## Outlook and Vision for Next Steps

### Neuromorphic Principles for Biomedical Applications

NPBH Workshop Chair, Ralph Etienne-Cummings, presented a summary of this meeting at the [NIH BRAIN NeuroAI Workshop](#) (33,34,78) where he identified four key technical advantages of neuromorphic principles for biomedical applications. These include power efficiency; on-chip, real-time processing; adaptability and resilience; and a naturalistic code for interfacing with the nervous system. Power efficiency — along with reduced power requirements for biomedical systems — lowers heat dissipation, which is particularly impactful for implantable devices. On-chip, real-time processing provides low-latency signal processing for closed-loop algorithms, avoids the complexity of transferring data to the cloud, and improves security by only storing data locally. Adaptability allows the system to learn and tailor the treatment paradigm as time progresses and the biological response changes. Lastly, the event-based spiking code of neuromorphic models and devices may enable improved neural interfaces by mimicking the natural spiking patterns of neuronal circuits in the targeted regions of the brain. Neuromorphic advancements may be especially important for clinical applications (see Figure 1) where a large amount of data is collected, predictive algorithms are required, and low heat-dissipation and extended battery-life are critical for implantable systems.

Workshop participants suggested that when applying neuromorphic principles to solve biomedical problems, the focus should be on the clinical application (which can often be neglected). Participants discussed the advantages in neural engineering of developing neuromorphic solutions that precisely target specific neural circuits for a disease rather than broadly stimulating brain regions. On the other hand, participants noted that engineers should be unrestrained to determine the most appropriate technical solution for the problem at hand. Panelists emphasized identifying specific applications where neuromorphic provides clear advantages (e.g., low-latency to achieve real-time closed-loop systems; extremely efficient and resilient circuits).

### Future Research Areas

*Biointerfaces*

#### Multimodal Biosensors

Participants discussed the need for better wearable sensors that move beyond the limitations of the clinic and collect data while subjects are mobile and engaged in natural behaviors. Several clinicians emphasized the need for technologies that monitor multiple biomarkers. This includes chemical signals, such as potassium, oxygen, sodium, and various neurotransmitters (up to 24 as described in (79)). Participants also discussed the need for multimodal recording devices that can simultaneously measure electrical activity alongside chemical signals to provide a more complete picture of neural function.

Important device specifications were discussed including biocompatibility, low immune response, stretchability, flexibility, adhesiveness, and the ability to either dissolve or remain stable in the body (e.g., heart vs. stomach) depending on the biomedical application. The potential for neuromorphic approaches to enable better sensor integration across modalities (i.e., molecular, pressure, electrophysiology) and to operate at different time-scales with low power consumption and minimal latency requirements was considered a strength.

Novel Materials

There were several presentations on novel flexible and stretchable materials, and there was consensus that these characteristics are a requirement for next-generation neuromorphic biointerfaces. Bao highlighted how the field is applying material science to replicate rigid silicon circuits in soft, embedded jelly-like materials, often with comparable electronic properties. Bao then presented a nanoconfined polymer semiconductor (DPPT-TT) in a matrix material (SEBS) that can be tailored to be stretchable, self-healing, or biodegradable (80,81). Wang also presented a soft, flexible, immune-compatible organic electrochemical transistor (OECT) using a semiconductor polymer (p(g2T-T)). Wang demonstrated how the foreign body response can be minimized by using backbone (selenophene) and side-chain (THP and TMO groups) molecular design strategies (82). Akinwande suggested that the human body should be integrated as an active component of the device rather than a passive target for treatment and observation. His work implemented ultra-thin graphene in electronic tattoos embedded in the surface of the skin (65). Panelists also discussed the importance of biocompatibility and CMOS integration with new materials.

Difficulties in clinical translation of novel materials were noted by participants. NIBIB's [Biomaterials Network](https://videocast.nih.gov/watch=45769) (https://videocast.nih.gov/watch=45769, starting at time 2:27:10) is an example of a complementary funding mechanism for novel neuromorphic materials. The biomaterials translation process is often encumbered by the so called "valley of death", beyond which few biomaterial technologies make the transition from bench to bedside. The Biomaterials Network aims to accelerate biomaterials-based technology development via a coordination center and translational resources that support technology development projects. The goal of this infrastructure is to facilitate collaboration, improve dissemination, and propel clinical translation and commercialization of biomaterials.

*Neuromorphic Neuromodulation*

Duygu Kuzum concluded that neuromorphics can help neuroscience and neuromedicine in two major ways: (1) neuromorphic computational models can integrate information from multiple modalities to create accurate, low-dimensional neural representations to improve neural prosthetics or BMI performance; (2) neuromorphic hardware can transform neural recording methodologies by enabling on-chip, real-time signal processing with learning capabilities. These capabilities can shift the processing location of the data from off-chip to on-chip and increase the longevity and stability of neural interfaces. Participants expressed that external reliance on internet resources also poses security risks, which can be obviated by on-chip processing. Participants asserted that neuromorphic neuromodulation devices can be developed using legacy process nodes (e.g., 180 nm, 65 nm) that would not depend on access to cost-prohibitive new infrastructure. One example was proposed by participant Tobi Delbruck who described an approach to short-term learning via the use of hidden persistent states that stores memories for arbitrarily long times using context-dependent control of a forgetting time-constant, which allows rapid memory updates (83). Delbruck contended that it was feasible to build on-chip mixed-signal CMOS controllers using affordable 65–nm technology that run on 1 V supply, utilizing 1 mm$^2$ of silicon and a 2-layer recurrent neural networks comprising 50k weights. This system would consume fewer than 5 μW (at 100 Hz update rate with a 250 kHz clock including a non-optimized low-power static random-access memory) and could power many real-time detection and control tasks in standard iEEG electrode arrays, especially for use in tandem with optogenetic stimulation.

*Real-Time, On-Chip Learning*

Participants highlighted that BMI signals are known to drift over time (84), suggesting that on-chip learning could be a fast and efficient way to correct signal drift regardless of the source of the drift. Participants also noted that interpersonal and intrapersonal variability in biological recordings could also be addressed by on-chip learning. This would benefit long-term experiments that track the same type of biological dynamics to study behavior and cognition in both animals and humans. An advantage of the real-time component is a reduction in processing time, as there is no longer a wait to transfer the data to the cloud. This could minimize delays in closed-loop algorithms. As decoding algorithms utilizing machine learning and spiking neural networks become more prevalent (85–87), participants concluded that shifting from cloud streaming to real-time, on-chip hardware processing (25) can be complementary and expand the technical capabilities of neuromodulation, BMI, and closed-loop biomedical systems.

*Next-Generation Prostheses*

Participants enthusiastically noted that neuromorphic hardware and principles appear ripe for prosthetic applications that require neural recording, decoding, encoding, and closed-loop control to enhance complex bidirectional sensory information processing. Panelists highlighted the potential for high-density neural probes to simultaneously record and stimulate, emphasizing that neuromorphic approaches excel at extracting sparse information for real-time, low-latency feedback systems. Preliminary payoffs are notable in visual prosthetics where the field can leverage extensive knowledge of the functional anatomy of the visual cortex and apply dimensionality reduction methods to predict the locations at which phosphenes will appear instead of reliance on patient reports (88). Participants argued that the peripheral nervous system could stand to benefit from neuromorphic strategies but is an "ignored cousin" despite its critical importance to brain research. There was general concurrence that neuromorphics is appropriate for the replacement of lost function. This has been observed in the history of cochlear implants, in addition to emerging evidence in sight restoration and somatosensory neuroprostheses (89).

*Time-Synchronized Multimodal Datasets*

The NIH BRAIN BBQS (Brain Behavior Quantification and Synchronization) program ([https://brain-bbqs.org](https://brain-bbqs.org)) — which aims to capture synchronized neural and behavioral data in naturalistic environments across both clinical and non-clinical populations — was mentioned by multiple participants. The importance of simultaneous, multimodal recordings with different sensor types (e.g., EEG, electrooculogram (EOG), electromyography (EMG), electrocorticography (ECoG)) capturing the same events and behaviors was emphasized. This neural-behavioral synchronization would allow for correlation across modalities and enable unconstrained algorithm research and development. Andreas Andreou noted that researchers such as Parisa Rashidi are collecting data from Intensive Care Units (ICUs) potentially with event-based cameras that could be of great value to the neuromorphic community (90). Andreou further noted the importance of looking beyond the neuromorphic world to the broader community for datasets. Sydney Cash noted that substantial intracranial recording datasets already exist through BRAIN Initiative data repositories like Data Archive for the BRAIN Initiative (DABI; https://dabi.loni.usc.edu) and the Distributed Archives for Neurophysiology Data Integration (DANDI; https://dandiarchive.org). These datasets have been previously leveraged in algorithm development competitions to solve biomedical problems such as seizure prediction. Multiple participants suggested that future algorithmic or data challenges using these multimodal datasets could feature neuromorphic algorithms, hardware, or codesign, and evaluate specific advantages over conventional approaches.

*Standards*

Participants expressed that the importance of open standards cannot be overlooked. It was noted that successful open standard models like USB and OpenGL transformed computing through common interfaces. The OpenGPU Foundation highlighted the neuromorphic intermediate representation (NIR (10)), a common tool-chain layer that translates between frontend simulated applications and neuromorphic modeling software and backend digital hardware platforms. NIR cleverly decouples the software from the hardware layer, enabling the potential for seamless integration with mixed-signal chips, analog hardware, and hybrid digital-analog systems. Thus far, NIR has successfully evaluated four hardware platforms and seven simulators. NIR represents an exciting direction for the field to promote wider adoption and faster development of neuromorphic systems.

*Benchmarks*

The value of benchmarking for advancing technology development was debated by participants: while common metrics for head-to-head comparison of different approaches enables objective assessment, some participants expressed concerns that such metrics can lead to incremental progress at the expense of potentially transformative innovation. Throughout the cycle of neuromorphic development, funders need to make evidence-based decisions and compare novel approaches to established benchmarks. These cycles may be different depending on the neuromorphic platform (e.g., chip-less wearables vs. on-chip CMOS devices). For example, NeuroBench (76) is a collaboratively designed framework for neuromorphic benchmarks that resulted from open community input across academia and industry, supported by cumulative efforts of the NSF-funded Telluride Neuromorphic Cognition Engineering Workshop, the Neural Inspired Computation Element (NICE) conference, and recently promoted at the International Conference on Neuromorphic Systems (ICONS) in 2025. NeuroBench provides flexible benchmarking that is intended to co-evolve with the field as it grows through competitions, workshops, and centralized leaderboards. In its current form, NeuroBench has two tracks: algorithmic and systems. The algorithm track is hardware-independent, utilizing dataset, algorithm, and metrics, while the system track provides end-to-end evaluation of an integrated system. NeuroBench has supported a community-driven Grand Challenge on neural decoding for motor control of non-human primates in IEEE BioCAS conferences in 2024 (91) and 2025. Benchmarking tools like NeuroBench may prove to be key enablers of community-driven advances for the neuromorphic biomedical field. Other important considerations for datasets and benchmarks described by the embodied neuromorphic intelligence community (see Box 6 in (92)) will also need to be considered given the inextricable link between embodiment and biomedical engineering.

*Infrastructure to Advance Neuromorphic Engineering*

The importance of temporal processing and the need to invest in research and design methodologies — like co-design algorithms with synchronous circuits or event-based (asynchronous) circuits — were stressed by participants, noting that current design automation tools cannot support automated workflows from algorithms to chip design. Neural chiplets (93) were posed as a way forward given their modularity, scalability, and reusability. The National Institute of Standards and Technology (NIST) National Advanced Packaging Manufacturing Program (94), initiated as part of the U.S. CHIPS and Science Act of 2022, was highlighted as a newly released funding opportunity relevant to the neuromorphic community; it was noted that biomedical packaging is within the scope of this Program.

The Neuromorphic Commons - THOR project, awarded by NSF in 2024 (Award No. 2346527, https://www.neuromorphiccommons.com), aims to develop and deploy large-scale neuromorphic computing research infrastructure offering community access to heterogeneous neuromorphic computing hardware systems through close-knit partnerships with industry. THOR offers (1) remote access to large-scale neuromorphic systems; (2) open-source hardware/software co-design frameworks and tools; (3) common benchmarks and competitions; and (4) rapid algorithm development by providing access to a collection of learning modules, network models, and example frameworks.

In addition, the NSF has sponsored the ACCESS (Advanced Cyberinfrastructure Coordination Ecosystem: Services & Support) Program, which provides access to graduate students, researchers, and educators in the United States to utilize advanced computing systems and services at no cost. For details about the program and eligibility, see https://access-ci.org.

*Computational Infrastructure to Improve Clinical Care*

Viktor Jirsa's presentation showed that virtual brain models derived from patient-specific datasets can demonstrate surprisingly accurate predictions for individual responses to different stimulation parameters and for determining the locations of epileptogenic zones outside areas accessible by implanted electrodes (44,45). Simulation-based inference for personalization requires thousands of model simulations, which presents a substantial computational bottleneck (e.g., Jirsa commented that 20 minutes of fMRI resting-state recordings could require more than two days of processing time even with Graphics Processing Unit (GPU) acceleration). These virtual brain models have not been implemented on current neuromorphic architectures such as SpiNNaker or BrainScaleS. It remains an open question whether other neuromorphic architectures — such as Intel's Loihi platform or other spiking neural network chips; for reviews of current and emerging platforms, see (9,95,96) — could address the computational bottleneck of brain-scale predictive modeling for epilepsy and other neurological diseases.

*New Approach Methodologies (NAMs)*

New Approach Methodologies (NAMs) are intended to accurately model human biology, encompassing *in vitro* organ-on-a-chip systems and *in silico* approaches performed by computing platforms or custom hardware, including mathematical modeling, simulation, and ML/AI approaches (97). Computational models (e.g., virtual brain twins (47)), neuromorphic emulators (e.g., neuromorphic twin (57)), and analog or mixed-signal neuromorphic devices may provide a basis for future energy-efficient, personalized, and safe NAMs with the potential to accelerate neuroscience and brain-health technologies. NAMs were not explicitly discussed at the NPBH workshop, but several related techniques were presented, including Jirsa's virtual brain twins and Donati's cortical emulator. The NIH recently (29 April, 2025) announced that it is prioritizing NAMs-based approaches: https://www.nih.gov/news-events/news-releases/nih-prioritize-human-based-research-technologies.

## Transdisciplinary Education

Based on community input, targeted educational initiatives to bolster neuromorphic engineering in neuroscience and healthcare may require building transdisciplinary teams with a shared vocabulary, complementary expertise, and accessible training that spans the research–translation pipeline. In the few postgraduate neuromorphic engineering programs that exist, training is heavily weighted toward physics, material science, and electrical and computer engineering (98). Many neuromorphic device engineers do not have formal training in neuroanatomy and neurophysiology, *in vitro* or *in vivo* neuroscience methods, clinical neuroscience, patient-centered design, the regulatory process, and/or neuroethical considerations for medical device development (98,99). For example, as of 2020, there were no specific postgraduate engineering programs that trained students in the field of implantable and wearable electronics for sensing and treating diagnosed neurological diseases (98,99). Therefore, neuromorphic engineering teams may not align technical capabilities, such as designing interfaces and algorithms, with real-world clinical impact.

Developing human- and brain-machine interfaces requires expertise and training in micro-electrode design, biocompatible materials, low-power circuit design, and real-time neural signal processing. For example, an emerging theme from the workshop, discussed by Duygu Kuzum and others, was the call to shift processing from off-chip, cloud-based computing to on-chip, edge computing in human- and brain-machine interface development (see *Engineering Challenges, Topic 4 - Medical Imaging, Wearables and Analysis, Day 2: Session 3 Materials for Neuromorphics (Devices)*). Training in the recording and analysis of high-dimensional, multimodal data from BMI recordings (see *Clinical Challenges*, *Data collection limitations*) was also identified as an important challenge.

On the clinical side, some attendees expressed the view that clinicians working with BMIs may have limited familiarity with neuromorphic platforms or hardware/software co-design principles. This knowledge gap may potentially hinder clinical input into the development of neurobiologically-inspired architectures and device design, which is critical to ensuring responsiveness to patient needs (see *Clinical Challenges*). A similar neuromorphic knowledge gap exists in traditional computational neuroscience training, which typically focuses on theoretical modeling and analysis, with minimal emphasis on the engineering demands of deploying algorithms in real-world clinical settings in medical devices or robotic platforms. Brad Aimone discussed how neuromorphic computing can help us understand neuroscience better by moving beyond traditional sequential processing models to more brain-like computational frameworks, including those incorporating heterogeneity, parallel sampling and spatial temporal processing (see Day 2 Session 4: Medical Imaging, Wearables and Analysis).

Future directions to address these knowledge gaps and challenges could include several training and educational activities to foster collaboration and understanding of neuromorphic approaches. These activities could focus on training clinicians and neuroscientists in neuromorphic approaches, while training material scientists and neuromorphic engineers in neuroscience. Such cross-disciplinary approaches have been employed in biophysics, which began by applying quantitative methods to biological problems (100,101). The training and educational activities illustrated below have been supported by U.S. funding agencies and could inform future transdisciplinary training opportunities for clinicians, computational neuroscientists, material scientists, and engineers.

*(1) Interdisciplinary Summer Schools, Hackathons, and Hands-on Workshops*

The NSF-funded Telluride Neuromorphic Cognition Engineering Workshop, Neuro-Inspired Computational Elements (NICE) program (see Benchmarks section), and the International Conference on Neuromorphic Systems (ICONS) tutorials are examples of joint community meetings led by transdisciplinary experts in the neuromorphic space. These workshops can range from a 1 day overview (e.g. NICE (https://niceworkshop.org) and ICONS (https://iconsneuromorphic.cc) ) to a 3 week intensive deep dive (e.g. Telluride (https://sites.google.com/view/telluride-2026/home)). Topics over the years across these three workshops have ranged from "neuroscience basics" for engineers (e.g., synapses, dendrites, travelling waves, oscillations, spatial navigation, auditory pathways, decoding speech, music perception, microsaccades, neurotransmitters, synaptic plasticity, neuromodulation, biological neural networks, sensory-motor systems, reinforcement learning, and decision making, etc.) to more focused neuromorphic computational hands-on workshop topics (e.g., NeuroBench, Nengo, BrainScaleS, SpiNNaker2, spiking neural networks on FPGA for embedded systems, neurorobotics, dynamic vision sensors, on-chip learning for signal-drift correction, multimodal biosensor integration, and real-time closed-loop control strategies).

*(2) Algorithm and Dataset Challenges*

During the discussion on Benchmarks and Data, several participants voiced the advantages of competitive challenges for advancing neuromorphic algorithms and datasets — a so-called "battle of the algorithms". The discussion included utilizing multimodal datasets to develop new neuromorphic algorithms and hardware, which could then be compared to conventional approaches to highlight the neuromorphic advantage. As referenced above (see *Time Synchronized Multimodal Datasets*), a wide variety of neural and behavioral multimodal datasets, including intracranial recordings in humans, are currently available through NIH BRAIN Initiative data archives such as DABI and DANDI, which could be incorporated into common benchmarks for challenges.

Challenges using data science methods could build a better understanding of high-dimensional multimodal neural and behavioral datasets by comparing different algorithms and their performance. During the workshop, it was noted that the American Epilepsy Society hosted an algorithm development challenge for seizure prediction through Kaggle (https://www.kaggle.com/c/seizure-prediction). These types of challenges encourage entries from transdisciplinary teams and incorporate benchmarks that consider both computational and clinically-relevant constraints.

*(3) Clinical Immersion Programs for Engineers and Material Scientists*

Clinical immersion programs for biomedical engineers currently exist at the undergraduate and graduate level and are funded by the NIH (102). The purpose of these programs is to expose biomedical engineers to clinical operations and medical work culture, with the end goal of identifying clinical challenges. This model could be tailored for neuromorphic engineers and materials scientists to identify unmet biomedical needs. Clinical settings for potential short-term rotations could include epilepsy monitoring units, rehabilitation clinics, medical imaging facilities, and prosthetic fitting centers. Potential areas of interest could include (but are not limited to) patient workflows, immuno-compatible bioelectronic devices, challenges in medical image reconstruction, and human factors in device use.

*(4) Materials–Device–Algorithm Co-Design Workshops for Biomedical Engineering*

Participants noted that there is a lack of datasets to facilitate co-design in neuromorphic materials, devices, and algorithms. Workshops have been held to address this co-design gap for neuromorphic hardware and software (examples include, but are not limited to: "NSF Workshop on Machine Learning Hardware Breakthroughs Towards Green AI and Ubiquitous On-Device Intelligence" (https://sites.google.com/rice.edu/ml-hw-greenai-on-deviceai/home), the "Workshop on Neuromorphic Computing: from Material to Algorithm (NeuMA) (https://sites.google.com/view/nuema/home), and the NSF Workshop on AI and HW 2035: Shaping the Next Decade (https://publish.illinois.edu/ai-hw-workshop/). However, there is a lack of focus on biomedical applications. Development of workshops that bring together materials scientists, chip designers, and algorithm developers from all sectors and engaging teams to work through integrated design problems — such as biocompatible stretchable interfaces with on-chip spike sorting, synaptic transistors, or memristor-based neuromorphic hardware vs. conventional deep neural network systems — is one approach to facilitate transdisciplinary teamwork. It was also noted during the workshop that material scientists are not well represented in the neuromorphic community, which is a missed opportunity to incorporate their expertise in the co-design process.

Continued support of these types of training and education activities would foster an understanding of neuromorphic methods, jump-start cross-disciplinary collaborations, and encourage the development of a shared vocabulary for the next generation of neuromorphic researchers.

# Conclusion:

The NPBH Workshop successfully brought together engineers, scientists, clinicians, and funders to explore how neuromorphic engineering can advance medical applications. The workshop included sessions on neuromodulation, prosthetics, devices, and wearables. Participants emphasized that when developing neuromorphic technologies, the focus should be on solving real world, clinical problems. The key is identifying specific applications where neuromorphic approaches provide clear, demonstrable advantages, particularly for real-time, closed-loop systems that require low latency and highly efficient, resilient circuits. Participants identified several significant challenges that the neuromorphic biomedical field faces, which were in the areas of clinical application, engineering, infrastructure, and commercialization. Throughout the workshop, participants also emphasized the need for transdisciplinary communication and education between engineers, neuroscientists, material scientists, and clinicians. This workshop has highlighted research areas of interest and identified existing federally funded programs that can be leveraged to further advance the field.

# Appendix 1: Meeting Synopsis

The two-day NPBH workshop was a single-track meeting structured to progress from cortical neuromodulation approaches to peripheral technologies and prosthetics, followed by a moderated discussion to generate a roadmap, and a funders panel to discuss regulatory considerations and funding opportunities. Day two focused on advances in materials research and enabling technologies, culminating in a moderated discussion with four themes: (1) benchmarks and data, (2) clinical need: sensor technology, device specification, data collection, (3) regulatory considerations and barriers, and (4) infrastructure needs.

## Day 1 - Opening Remarks

Ralph Etienne-Cummings introduced the concept of neuromorphic principles by drawing inspiration from nature with examples from Antoni Gaudi (Nautilus shell for architectural designs), Leonardo da Vinci (sycamore seed-inspired helicopter aerodynamics), John Hopfield (associative memories), Paul Mueller and Carver Mead (transistor-based neuron models (4,5)), and Misha Mahowald and Eric Vittoz (nature-inspired electronics). He presented Gert Cauwenberghs' neuromorphic analysis by synthesis framework (103) highlighting the different modeling granularities across many spatial scales, from single-channel transistor models (the way transistors operate in the subthreshold) to whole-brain computational processes, emphasizing the importance of considering the entire continuum. Ralph asserted that neuromorphic principles can impact healthcare in multiple domains including diagnosis, neural interfaces, biological analysis, and the development of specialized neuromorphic tools for medical applications. He concluded by reminding the audience that the definition of "neuromorphic" has evolved from strictly hardware-specific implementations to encompass low-level sensory processing, AI algorithms, data science approaches, and computational acceleration for healthcare applications.

Andrea Beckel-Mitchener, Deputy Director of the NIH BRAIN Initiative, explained that BRAIN is a global effort focused on discovery and technology development for better understanding the human brain. Since 2014, BRAIN has invested over 3.5 billion dollars in hundreds of projects, yielding remarkable discoveries about brain function. Building on these significant achievements, BRAIN is interested in leveraging knowledge to explore new areas such as neuromorphic engineering and systems based on neuromorphic principles as potentially transformative approaches. This emerging field promises new architectures, energy-efficient computing devices, and novel algorithms poised to enhance our understanding of the brain, while also contributing to knowledge about health and disease and innovative biomedical device designs. She joked "I know we're focused on biomedical applications here, however, we might just save the planet along the way…". BRAIN maintains an assertive data sharing policy (104), making wiring diagrams of multiple organisms' brains, functional electrophysiological measurements, and cellular "parts lists" available through NIH-supported archives for researchers to access and utilize and encouraged their use of BRAIN knowledge bases. A distinctive feature of BRAIN lies in its interdisciplinary approach, recruiting not just traditional NIH grantees like biologists, but also researchers from physical sciences and engineering, which has proven extremely fruitful for developing new technologies and approaches.

Paul Lane, Acting Deputy Division Director, NSF ENG Division of Electrical, Communications, and Cyber Systems (ECCS) Division started his welcoming remarks with a famous line from The Graduate with a slight update: "Our protagonist, Benjamin, is at a party with a bunch of older individuals. One man comes up to him and talks about the future. He says one word: plastics. We're here to update that. I might say one word: energy". He asserted that the transformative advances in society, in medicine, from artificial intelligence, generative AI, to large language models all require enormous amounts of energy. And so applying neuromorphic principles to new circuit designs, devices and materials from the bottom up will be essential. He has been interested in neuromorphics since 2018, participated in the Brain Inspired Dynamics for Engineering Energy-efficient ([BRAID](#)) (105) Emerging Frontiers in Research and Innovation (EFRI) topic, and foresaw neuromorphic in the Future of Semiconductors ([FuSe2](#)) (106) programs in terms of spintronics, memristors, phase change materials, novel gating concepts, ion gating, even string gating. He hoped to see new partnerships and ideas emerge from this workshop.

David Rampulla, Deputy Director of the Extramural Science Program, Acting Director of the Division of Discovery Science and Technology (Bioengineering) at NIBIB, viewed neuromorphic engineering and neuromorphic computing as a window to next generation biomedical technologies, backreferencing Ralph's opening remarks that highlighted benefits of neuromorphic technologies to the whole body. He echoed that these benefits include power consumption, health, and privacy highlighting the value of on-chip computing that negates the need to send data to power-hungry data farms. He concluded with the observation that neuromorphic computing offers the possibility of communicating directly with human biology which has the potential to open up new ways of engineering biology for human health.

## Day 1 Keynote - Towards "Neuro/Physio-morphic" Devices [Generation 3?]

Timothy Denison, Professor of Engineering Science, University of Oxford, presented his vision for next generation adaptive DBS devices starting with a retrospective on Medtronics's Percept$^{TM}$ for which a clinical trial was recently completed. The journey from open-loop to closed-loop neuromodulation is significant; he emphasized the importance of not just neuromorphic but also closed-loop "physiomorphic" approaches using physiological markers to adjust stimulation in real-time to better match brain states. A seminal paper by Peter Brown and Simon Little (107) long established a correlation between beta-band power and symptoms in Parkinson's patients, enabling a thermostat-like approach where stimulation would be activated when power exceeded a pre-set threshold. However, most adaptive stimulation research was conducted between "noon and teatime," raising questions about how these systems perform across different times in the circadian cycle, especially during sleep. Future devices should thus incorporate both reactive and predictive capabilities, similar to how bees predict seasonal changes rather than just reacting to immediate conditions, creating a truly "physiomorphic" system.

Developing adaptive devices requires balancing clinical necessity with practical elements such as economic viability, workflow considerations, and navigating complex regulatory pathways. A major challenge in developing novel bioelectronic devices is the "chicken and egg" problem - needing data to justify investment but requiring investment to gather the necessary data. Denison advocated for an incremental approach using "science platforms" that build upon existing technologies with guaranteed therapies while enabling exploration of novel concepts with minimal additional investment, highlighting a collaboration that entailed just one firmware engineering (<$1M dollars) on spinal cord stimulation using neuromorphic processing (108). Denison is a proponent of figuring out what one can do to incrementally improve existing technologies given the "hidden capabilities" in them.

Denison reminisced about NeuroPace's Responsive Neurostimulation (RNS) while it was still in development, reminding that audience that early stimulators were conceptualized as "reversible lesions", which may have anchored the field to static approaches rather than dynamic physiological interventions. A key design constraint is power consumption. Stimulators typically run at around 100 microwatts (while cardia pacemakers run around 1 microwatt) requiring adaptive sensing and control systems to operate at about one-tenth that power level. Trained as a circuit designer and influenced by many hours of ultra-low power design conversations with Rahul Sarpeshkar (109), Denison developed a "brain radio" (110) to extract microvolt-level signals from the basal ganglia while consuming only 5 microwatts, with another 5–10 microwatts for the algorithm to adjust stimulation. Other early successes included using cortical sensing for brain-computer interfaces in ALS patients (111,112) and essential tremor (113), but challenges remain in sensing from deeper brain structures like the basal ganglia and thalamus where signals were 10–20 times smaller than seizures.

The BRAIN-funded Summit RC+S device (39) represented a technological advancement, moving from analog preprocessing to on-chip fast Fourier transforms that reduced power consumption to nanowatts per channel while enabling measurement across multiple frequency bands. This system allowed one to set lower and upper bounds and an in-range "no action required" bound, similar to the theory of Walter Cannon to adhere to how human physiology works. At issue is risk management and ways for regulators to ensure that neuromorphic or physiomorphic technologies would be safe. (Interested readers are referred to Guidance document 60601-1-10 (72) and Appendix 1 - Day 1 Meet the Funders.) Denison clarified that Summit RC+S device was embedded with actuation limits, upper and lower bounds, and fallback mode to ensure safety. "With these safety systems in place, it created a sandbox where we could explore new algorithms." Denison suggested that a great way to engage the community on novel algorithm design could be through "the battle of the algorithms" which would allow different devices to run different algorithms for different diseases and allow researchers to monetize their ideas.

The ongoing ADAPT-PD study aimed at demonstrating safety and effectiveness of adaptive DBS algorithms in Parkinson's disease patients (Medtronic Percept$^{TM}$ device) (37,38) compared two control algorithms: a single-threshold "bang-bang" approach versus a dual-threshold adaptive DBS homeostatic approach inspired by physiological systems that allow parameters to vary within acceptable limits. Study results showed that the dual-threshold homeostatic approach demonstrated statistically significant clinical improvements over the single-threshold approach, suggesting that mimicking natural physiological responses may be more effective. Denison noted that 30-60% of the variance in the beta band for PD patients is related to time of day, suggesting future devices should incorporate circadian rhythmicity into their design. In conclusion, energy efficiency remains a critical consideration, with neuromorphic approaches needing to demonstrate advantages over classic digital signal processing, particularly for implantable medical devices. Future challenges include designing for manufacturability, test reliability, managing signal drift in analog systems, system verification, risk management, and optimizing across multiple dimensions rather than focusing on a single parameter like power consumption.

### Day 1 Keynote – Q&A Highlights

Sydney Cash from Massachusetts General Hospital inquired about the on-chip versus off-chip processing dichotomy, noting the tension between miniaturizing computing power onto chips versus streaming data to perform processing externally in the cloud. Denison acknowledged system distribution as an important design consideration, referencing his work with Cortec's brain interchange system which streams data out for external processing before sending commands back, with latency being a critical factor. Denison noted a multi-tiered approach demonstrated by Greg Worrell at Mayo Clinic using Summit RC+S, employing simple embedded algorithms for quick responses while simultaneously streaming data to patient-carried devices and cloud processing for more sophisticated algorithms. He emphasized that patient experience must be considered when designing distributed systems, noting practical challenges such as radio links breaking when patients roll over during sleep and the burden of requiring patients to manage multiple external components.

Duygu Kuzum from UCSD questioned the reliance on local field potentials (LFPs), highlighting their complex nature as vaguely defined spatiotemporal averages whose information content depends heavily on measurement methods, particularly with large DBS electrodes that introduce significant spatial averaging. Denison clarified that while LFPs have limitations, they have proven reliability over long-term recordings compared to alternatives, acknowledging that they remain susceptible to artifacts such as ECG signals that can be picked up when devices are placed in the pectoral region.

In response to Tobi Delbruck's online question about epileptic seizure detection and quenching, Denison suggested the field needs to regroup with fresh perspectives on epilepsy treatment, particularly questioning whether current approaches like RNS might be triggering interventions too late in seizure progression. Denison critiqued current approaches including the SANTE trial's application of vagal nerve stimulation parameters to new networks and disease states, noting unintended consequences such as sleep state corruption when stimulating the anterior nucleus of the thalamus during nighttime.

Sihong Wang from the University of Chicago asked whether future neuromorphic analysis circuits should wirelessly communicate with sensors and stimulators or be implanted together as an integrated system. Denison emphasized that radio telemetry remains relatively expensive compared to on-chip computation, with much of his current DyNeuMo (114) study focused on practical issues like teaching families about recharge management rather than algorithms or neuroscience principles. He preferred incorporating more processing within implanted devices given current limitations, noting that telemetry typically consumes an order of magnitude more power than on-chip algorithms. This was confirmed by Ralph Etienne-Cummings who clarified that published literature showed telemetry used approximately 80% of on-chip power consumption (74,75).

## Day 1 Session 1: Epilepsy & Cortical Disorders (Neuromodulation)

Session 1, moderated by Grace Hwang of NINDS and Jessica Falcone of NIBIB, was a deep dive into epilepsy diagnosis and care using DBS and recording technologies and the virtual brain.

The first speaker was William Stacey, M.D., Ph.D., of the University of Michigan. His talk, *Data Driven classification without a gold standard: epileptic HFOs*, introduced that epilepsy affects one in a hundred people with one-third not responding to treatment which presents substantial challenges in understanding seizure origins and developing effective treatments. The current surgical decisions for intractable epilepsy rely heavily on empirical observations using visual analysis of EEG patterns derived from decades-old methodologies. Existing neuromodulation devices (VNS, DBS, RNS) were never properly optimized for epilepsy treatment, with stimulation parameters often borrowed from other applications. Some closed-loop devices sense and stimulate with four-year battery life but suffer from extremely low processing power: "24 hours of recording results in just five minutes of saved data and only from a few channels. The problem is that the state of the art for clinicians is just not good enough. We keep inventing and adding new things to the clinical pipeline. And it really hasn't made much of a difference." He investigates high frequency oscillations (HFOs) which represent a promising biomarker for identifying epileptogenic zones, but distinguishing pathological from normal HFOs remains challenging due to lack of gold standard validation. Advanced data-driven approaches, including non-negative matrix factorization, graph theory, one-class support vector machines and Uniform Manifold Approximation and Projection (UMAP) clustering of millions of HFOs show potential for identifying epileptogenic tissue with greater specificity [(41,42), and unpublished data]. A recent multi-center HFO study also showed promising results for detecting seizures and other intracranial EEG patterns to improve postsurgical seizure outcome (40). Neuromorphic is nascent in epilepsy, uses include real-time seizure prediction (IBM TrueNorth (48)) and real-time HFO detection from intracranial EEG (22,49). In conclusion, Stacey asserted that current implantable devices for epilepsy suffer from significant limitations in sampling frequency, bit precision, and computational capabilities, hampering research progress. The field needs new analytical tools to process high-resolution, high-throughput EEG data, and low-power devices to better interface with the brain, regardless of whether they specifically mimic brain structure.

The second speaker was Dr. Sridevi Sarma of Johns Hopkins University. Her talk, *A Resting State EEG Biomarker Points to Seizure Onset Zone*, focused on drug-resistant epilepsy patients whose surgical outcomes vary significantly, with success rates averaging around 50% for seizure freedom following resection. Her lab developed the "EZtrack-RS" software tool that uses resting state intracranial EEG to identify potential epileptogenic zones without relying on seizure data, functioning as a decision support tool. Her approach reframes the problem by focusing on the inhibition hypothesis: epileptogenic zones may be strong "sinks" being inhibited by "sources" in the brain's network during non-seizure states. Network nodes can be determined based on just 5 minutes of intracranial EEG data. She has tested EZtrack-RS on 65+ patients across six centers demonstrating that overlap between predicted epileptogenic zones and surgically treated areas correlated with successful surgical outcomes (43). She asserted that neuromorphic computing is relevant for real-time closed-loop control such as in 1) using a neuromorphic feedback controller for drug release to prevent seizures via a notional non-invasive ultrasound neuromodulation system, or 2) for medical applications where high spatial and temporal precision is needed to rapidly alter brain or peripheral nerve temperature using a thin-film thermoelectric (115) device. Sri explained that her Ph.D. training was in pure control theory; she became interested in neuroscience in graduate school when she studied with the late Suzanne Corkin on motor systems and PD. It is her view that advanced control theory applications in neuromodulation have not kept pace with developments in other fields, with most biological control systems still using simple threshold or proportional approaches. Modern control methods (optimal control, robust control, network control) require more sophisticated computational approaches that could benefit from neuromorphic implementation. She concluded that neuromorphic computing could provide the necessary computational capacity for implementing sophisticated feedback controllers in applications ranging from focused ultrasound to thermoelectric neuromodulation.

The final speaker was Dr. Viktor Jirsa of Inserm. He was heavily involved with the Human Brain Project that used neuromorphic architectures including SpiNNaker and BrainScales wafer. While the Human Brain Project just ended in 2023, the neuromorphic infrastructure remains active in the digital neuroscience infrastructure, EBRAINS (https://www.ebrains.eu/). Jirsa's talk, *Virtual Brain Twins for Medicine*

- *Encephalomorphic Systems*, featured digital twins using an individual's neuroimaging data (including diffusion tensor weighted imaging) to model brain dynamics at various scales, enabling prediction of epileptogenic zones and responses to neurostimulation. He explained that current virtual brain models operate with 200-300 nodes representing brain regions, but are scaling to 200,000 nodes to achieve millimeter-scale resolution needed for neurostimulation applications. Already virtual brain models have predicted epileptogenic zones that lie outside areas accessible by implanted electrodes (44–46), potentially explaining cases of unsuccessful surgical outcomes (29% of regions linked to failed cases). Digital twins built from patient-specific data show surprising predictive power for individual responses to different stimulation parameters in terms of spatial-temporal dynamics. However, major computational challenges exist in running high-resolution virtual brain simulations: 20 minutes of fMRI resting state recordings require more than two days of processing time even with GPU acceleration. Current neuromorphic architectures like SpiNNaker and BrainScaleS have not been successfully applied to these full-brain models due to limitations in handling time delays, noise, and high dimensionality. He concluded that simulation-based inference for personalization requires thousands of model simulations, creating a substantial computational bottleneck that neuromorphic architectures could potentially address.

## Day 1 Session 1: Panel Discussion Highlights

Three clinicians (Sydney Cash of Mass General, Nathan Crone of Johns Hopkins, Kendall Lee of Mayo Clinic) and two neural engineers (Giacomo Valle of Chalmers University of Technology in Sweden, and Chris Rozell of Georgia Tech) joined the Session 1 Panel discussion. When asked what they would do if given a low power implantable chip, most clinicians emphasized the need for devices with more channels, better sampling rates, and greater signal-to-noise ratios in fully implantable wireless form factors. Kendall Lee impressed upon the panel that the brain is electrochemical; neurochemical sensing presents promising opportunities alongside electrophysiological approaches but generates enormous amounts of data requiring advanced processing. Lee further mentioned multiple cyclic square wave voltammetry (MCSWV) and N-shaped multiple cyclic square wave voltammetry (N-MCSWV) that now allows monitoring of absolute value neurochemical changes in the brain that can be used for electrochemical closed-loop neuromodulation (116,117). Viktor Jirsa pointed out that multiple chips would be needed because the brain is a spatial temporal system. Nathon Crone agreed, explaining that there is a growing appreciation that epilepsy arises from broader brain networks than previously thoughts, and thus it is important to sample as much of the network as possible to develop effective ablative and neuromodulation therapies. The panel was in concurrence that time scales present a significant challenge in optimizing treatments, with months often needed to determine if interventions are effective, particularly in psychiatric applications. Collectively, the panelists identified a critical need for intermediate biomarkers that could provide earlier indications of treatment efficacy before clinical symptoms change. Sri Sarma highlighted the importance of maintaining space for scientific discovery alongside AI and neuromorphic approaches, as fundamental understanding of brain health is still lacking. The limitation of current data collection in epilepsy was also discussed, with high-resolution data primarily available only during inpatient monitoring rather than in ambulatory settings; datasets from ambulatory devices are extremely limited.

## Day 1 Session 2: Human-machine Interfaces (Prosthetics)

Session 2, moderated by Luke Osborn of Case Western Reserve University and Francisco Valero-Cuevas of the University of Southern California, focused on how neuromorphic principles and technologies may enhance HMIs, wearable sensors, and rehabilitation technologies.

The first speaker was Dr. James Cotton, a physician scientist at Shirley Ryan AbilityLab. His talk, *Clinical Gaps and Understanding Real World Outcomes for Prosthesis Users with AI Technologies*, focused on the gaps in understanding real-world function and how technology readiness levels help contextualize how prepared innovations are for actual implementation in patient care (118). He emphasized the crucial need for prosthetic devices to demonstrate improved real-world function through home trials rather than laboratory assessments, pointing out that lab-based evaluations often don't predict how technologies will perform in daily life situations (119). He explained that patients prioritize comfort, reduced pain, and psychological factors like avoiding embarrassment from falls over advanced technical features, with studies showing socket fit issues and skin problems are what patients most frequently report (50). His research focused on developing accessible movement analysis systems for clinical settings that can capture rich biomechanical data, which could provide better individualized understanding of patients' movement patterns for predicting fall risk and improving interventions (120). He proposed a framework using computational methods to link clinical assessments with real-world activities, creating "digital twins" that could simulate how different interventions might affect patient outcomes without requiring extensive randomized controlled trials. He concluded by emphasizing the importance of causal modeling approaches to understanding the complex interactions between brain signals, prosthetic devices, and rehabilitation outcomes for improving longitudinal patient data.

The second speaker was Dr. Ranu Jung, an Associate Vice Chancellor and Distinguished Professor at the University of Arkansas. Her talk, *Convergence: Deploying Neuroelectronic Interfaces*, emphasized that technology development must begin with the desired impact in mind, addressing not only the traditional "valley of death" between lab and commercialization, but also the gap between commercialization and widespread deployment which involves economics, policies, politics, and behavior. Jung explained that the healthcare spending trend is projected to increase by 748% for wellness technologies by 2040, while spending on disease treatment is decreasing, highlighting the importance of developing technologies that make economic sense and contribute to well-being rather than just treating disease. She asserted that the fundamental challenge of biohybrid systems, as described in her book (51) remains largely unsolved: biological systems are non-linear, non-stationary, multi-scale and plastic, while existing technologies, despite some adaptive algorithms, remain relatively fixed. Jung presented her work on respiratory pacing using neuromorphic controllers that incorporate adaptive pattern generators to respond to changing metabolic needs, such as $CO_2$ levels, automatically adjusting tidal volume and respiratory rate (59). This respiratory pacing system used a computational model with a neuromorphic closed-loop control system that can automatically personalize stimulation patterns and make real-time predictions. For prosthetic limbs, Jung described a fully implanted system developed since 2006 (as part of an NIH Bioengineering Research Partnership Program and DARPA's RE-NET and HAPTIX programs, and the Joint Warfighter Medical Research Program) that provided sensory feedback by directly stimulating nerves, with sensors in the prosthetic hand communicating wirelessly with implanted electrodes. The prosthetic system used a distributed interfascicular stimulation approach with longitudinal wires placed along nerves, allowing for selective stimulation by varying electrode location (52,53), configuration, and waveform shape. Jung emphasized that while researchers focus on specific sensory modalities like pressure or proprioception, patients may care more about other aspects of sensation that are practical and useful to them, such as simple tingling sensations with good clarity and location. She concluded her presentation with two case studies demonstrated profound impact: one participant seamlessly integrated the sensory feedback after years of non-sensorized prosthetic use as part of the NIH funded NEPH Study (121), while another experienced sensation in his missing hand for the first time in 74 years, described by his wife as a "miracle."

The third speaker was Dr. Elisa Donati of the University of Zurich and ETH Zurich. Her talk, *Neuromorphic Hardware for Closed-loop in Healthcare*, described neuromorphic hardware as an ideal approach to communicate natively with the human nervous system. It can address key challenges in neural interfaces by working in a low-power regime through event-based processing, which only processes relevant information rather than entire signals continuously. She argued that neuromorphic systems enable local processing of neural data, reducing bandwidth requirements for wireless transmission, while also allowing for real-time adaptation and learning with superior spatial-temporal resolution compared to conventional approaches. She presented her work on mixed-signal neuromorphic architectures that use analog subthreshold circuits to emulate neural behavior, with transistors functioning like "Lego blocks" that can be combined to create neurons and synapses with customizable features including a continuous prosthetic hand control that maps electromyography signals to finger movements. This system achieved comparable accuracy to deep learning approaches using significantly fewer parameters and lower power requirements ((56), Manneschi 2024 in prep). She described her neuromorphic approach for tactile feedback in prosthetics, showing how a simple spiking neural network with under 1,100 neurons could match the 90% classification accuracy of a CNN for identifying 17 different grasped objects (55). She also developed a "Neuromorphic Twin" with just 200 neurons to emulate cortical layer 4 of the somatosensory cortex that accurately replicated adaptation phenomena seen during invasive stimulation in human patients, where subjects stop feeling sensations after continuous stimulation (57). Donati concluded by applying neuromorphic technologies to create adaptive pacemaker based on central pattern generators for cardiac activity monitoring (58), long-term monitoring in patients with dementia focusing on multi-days and multi-subject adaptability, and post stroke rehabilitation.

## Day 1 Session 2: Panel Discussion Highlights

Xing Chen of the University of Pittsburgh, Nitish Thakor of Johns Hopkins University, Helen Huang of University of North Carolina, Vikash Gilja of the University of California, San Diego (and Paradromics), and Shih-Chii Liu of the University of Zurich joined the panel along with the speakers and session co-chairs. When asked about what impact neuromorphic principles could have in conveying highly complex information back to humans through user interfaces, Xing Chen framed the challenge in terms of efficacy, feasibility, and safety for visual prostheses. One needs a device that can be calibrated quickly during a clinical trial where it will not take months of manual reporting to determine phosphene locations. In terms of safety, one needs to ensure that stimulation does not lead to propagating activities and generate epileptic seizures. Francisco Valero-Cuevas emphasized focusing on enabling percepts rather than replicating signals, noting that biological systems naturally filter sensory information and adapt to create meaningful perceptions regardless of input modality. Nitish Thakor identified three key neural interface components (decoding, encoding, and control) acknowledging unresolved questions about modeling receptors in the periphery, and understanding nerve and fascicle encoding precision. Shih-Chii Liu highlighted the exciting potential of high-density neural probes (10,000 electrodes) for simultaneous recording and stimulation, emphasizing that neuromorphic approaches excel at extracting sparse information for real-time, low-latency closed-loop systems. She speculated that with 10,000 electrodes, one could imagine stimulating different layers of the cortex and how that could dramatically assist a blind individual one day.

When asked if low power system that adapts to evolve with users would benefit the user or rehabilitation, Helen Huang discussed the importance of personalization in medical devices, noting that intersubject variation requires adaptive approaches at different time scales, suggesting that reinforcement learning could continuously update device parameters. Helen also expressed enthusiasm for neuromorphic analog approaches owing to its latency, especially if these circuits could directly interface with the human nervous system. Multiple panelists addressed the challenge of sparse and variable data in neural interfaces, with Elisa Donati pointing out that signals change even within the same subject over time due to electrode shifts or cellular turnover. James Cotton suggested that correctly implemented AI approaches could help by aligning latent manifolds between individuals to enable more rapid calibration (referencing Lee Miller (122)), while Vikash Gilja described how speech restoration systems leverage foundational models from healthy individuals. Francisco Valero-Cuevas argued that neuromorphic/physiomorphic computing excels at embedded logic (contrasting with big data/optimization) because it can leverage insights from the physical substrate where learning occurs.

When asked about hardware and software limitations, Nitish Thakor noted significant challenges in conveying neuromorphic concepts to established neuroscience communities that are based on experimentally driven, mechanistic, reductionist neuroscience. Even the established neuromorphic DAVIS camera (https://inivation.com/solutions/cameras) is a small part of the larger camera ecosystem. Xing Chen highlighted preliminary payoffs in visual prosthetics, whereby carrying out recordings from the brain simultaneously during stimulation, the field is able to leverage extensive knowledge of the functional anatomy of the visual cortex and apply dimensionality reduction methods to predict the locations at which phosphenes are going to appear instead of reliance on patient reports. The panel concluded with calls for demonstrating clear clinical benefits of neuromorphic approaches, bridging gaps with traditional neuroscience, and developing multidisciplinary teams that include neuromorphic engineers, neuroscientists, and materials scientists to create better neural interfaces to communicate with the nervous system.

## Day 1: Moderated Discussion

This discussion was moderated by Sunny Bains of the University College London who began with exploring three potential areas where neuromorphic computing could contribute to biomedical applications: advancing scientific understanding of neural signals, developing implantable devices with low power requirements, and enabling personalization through adaptive circuits that can change as patients' needs evolve over time.

Grace Hwang raised a critical question about how neuromorphic approaches might address the problem of variability within individuals, where prosthetic training can become ineffective after just two weeks due to neural changes or electrode movement. Elisa Donati explained that despite changing brain recordings, stable "latent dynamics" can be identified in lower-dimensional space, and neuromorphic systems are particularly good at extracting these underlying dynamics and working with them. Viktor Jirsa introduced the concept of "degeneracy" as a fundamental challenge in neural systems, where the same individual can generate identical behaviors through different neural pathways, creating difficulties for both encoding and decoding neural signals. James Cotton noted that biological systems naturally have variability in dimensions that don't affect task performance, suggesting that effective assistive systems should mirror this flexibility rather than trying to eliminate all variability.

Elisa Donati asked why SpiNNaker and BrainScaleS were unable to support Viktor Jirsa's virtual brain twins to which Viktor clarified that these systems were optimized to use spiking neurons while the virtual brain operates at the mean field level giving rise to a fundamental mathematical incompatibility issue.

Ralph Etienne-Cummings emphasized that neuromorphic approaches seem particularly well-suited for prosthetic applications that require the full neural recording, decoding, encoding, and control loop, asking the panel whether this is where neuromorphic technology is most ready for deployment. Nitish Thakor highlighted the need for better tactile modeling across the entire neural chain from receptors, to fascicles, to cortex for sensory prosthetics highlighting Luke Osborn's neuromorphic prosthesis work (23). Thakor further noted that some models are ignored (123,124) while the peripheral nerves remain an "ignored cousin" compared to brain research despite their critical importance. Xing Chen enthusiastically expressed her interest in combining neuroregeneration therapies and pharmaceuticals with closed-loop stimulation in the peripheral or central nervous systems, referencing Gregoire Courtine's work in which chemical cocktails were injected into the spinal cord for closed-loop stimulation (125).

Sydney Cash proposed that while the interfaces with the nervous system need to be neuromorphic or physiologically compatible, the computational components in between could potentially use any technology that works effectively. Multiple participants debated whether the value of neuromorphic approaches should be evaluated based on practical performance metrics (size, weight, power, latency, throughput, efficiency) rather than philosophical adherence to biological principles. Roger Miller emphasized the importance of physiologically appropriate signals based on experience with cochlear implants. While cochlear implants work well when electrodes are properly positioned to replicate natural frequency mapping, misplaced electrodes (where frequencies don't match their normal locations) require patient adaptation that doesn't always succeed. So there really is a value in finding what the physiological signal is and using that to design a neuroprosthesis.

Chris Rozell observed that those working on replacement of lost function (e.g., prothesis, BMI, cochlear implant) seemed more enthusiastic about neuromorphic solutions than those focused on modulating aberrant neural activity for conditions like epilepsy or DBS for psychiatric neuromodulation. James Cotton countered that as a clinician he's agnostic if the solution is neuromorphic, he would prescribe a wheelchair in the foreseeable future because exoskeletons, though are more neuromorphic than wheelchairs, are far from usable for most patients.

Brad Aimone encouraged clinicians to "dream big" about what algorithms they would want in implants if power and latency constraints weren't an issue, suggesting neuromorphic hardware opens a wider aperture of computational possibilities. Kendall Lee described his neurosurgical perspective that the ideal neuromorphic solution would precisely target the specific neurons involved in a disease rather than broadly stimulating brain regions, comparing it to understanding the principles of flight rather than simply mimicking birds. Kendall further asserted that we can't change the brain but we can change the computer to better match how the brain works, arguing that it is better to interact with the brain through spikes. David Wyatt cautioned about dictating the engineering solution and suggested that it is best to focus on the problem and patient needs, and challenge engineers to propose solutions that better solve the problem, whether that is neuromorphic, quantum, analog, MIMD, von Neumann, or other. Tim Denison emphasized the importance of keeping a systems mindset and shared the story of how Medtronic revolutionized adaptive cardiac pacemaker using motion crystal with piezo electric to avoid heart failures.

The discussion concluded with Grace Hwang emphasizing the importance of identifying both short-term and long-term technological solutions for specific diseases and suggesting that neuromorphic approaches shouldn't be constrained to human biology but could draw inspiration from many species.

## Day 1: Meet the Funders

Dr. Zach McKinney of the United States Food and Drug Administration (FDA) provided a high-level overview of FDA's regulatory framework and programs for novel medical devices. He explained that the FDA regulates approximately 20% of the US economy, with the Center for Devices and Radiological Health (CDRH) specifically overseeing medical devices through a risk-based approach considering both intended use and technological characteristics. FDA uses a three-tier classification system: Class 1 (low risk) devices are generally exempt from pre-market review; Class 2 (moderate risk) typically follow the 510(k) pathway to market clearance via demonstration of substantial equivalence to one or more legally marketed "predicate" devices; and Class 3 (highest risk) require pre-market approval. The Q-submission program provides free opportunities for device developers to engage with FDA early and often, allowing for feedback on scientific approaches before formal submissions. Investigational Device Exemptions (IDEs) are required for significant risk clinical studies, with approval based primarily on patient safety, benefit/risk, and informed consent considerations rather than scientific merit. FDA recognizes three clinical study stages: early feasibility studies (small-scale prototype testing), traditional feasibility studies (pilot studies with final device), and pivotal investigations (final device for marketing application). The Breakthrough Devices Program expedites development and review for high-impact devices treating life-threatening or irreversibly debilitating conditions, providing faster review cycles and access to the Total Product Lifecycle Advisory Program (TAP). He encouraged developers of neuromorphic technologies to engage early and often through the Q-submission process, as regulatory approaches will be determined case-by-case based on comprehensive risk-benefit assessments in view of each device's technological characteristics and intended use. FDA has also been working to develop Guidance on "predetermined change control plans (PCCPs)" that (when finalized) (126) may provide a systematic mechanism to account for the adaptive nature of neuromorphic technologies in regulatory submissions.

David Wyatt, founder of the OpenGPU Foundation, described his motivation for funding neuromorphic standards and technologies by pointing to the unsustainable power consumption of current AI data centers, with individual server racks consuming 128 kilowatts each (equivalent to six homes) plus 60+ liters of water per minute; exponential growth in AI model complexity and computational demands, doubling approximately every 3.5 months (Moore's Law squared), creating unsustainable energy requirements that cannot be solved simply by expanding the power infrastructure; and biological systems that perform remarkable computations with minimal energy (typically about 20 watts), leveraging enormous interconnectivity and temporal processing that fundamentally differs from traditional von Neumann computing architectures. The OpenGPU Foundation is focused on funding test chips, especially neuromorphic exemplary applications that demonstrate advantages in temporal, stochastic, or multi-dimensional computing, and energy efficiency. OpenGPU aims to create standardized hardware-software interfaces to accelerate broader adoption by application and systems developers, as the OpenGL (Open Graphics Library) application programming interface (API) did for early GPUs and Nvidia's proprietary CUDA (Compute Unified Device Architecture) framework did for general-purpose computing on GPUs (GPGPU). Wyatt envisioned interfaces that could potentially extend to "wetware" (biological and hybrid) systems, and suggested this could all be achieved by following the successful model of nimble open standards associations—like USB, RISC-V, and OpenGL—through industry-wide groups which propose, develop, and maintain open protocols to enable vibrant and interoperable technology ecosystems.

Recent advances in neuromorphic computing include the development of neuromorphic intermediate representation (NIR,(10)) for interoperable software development across vastly different neuromorphic hardware processor architectures and efficient state space models based on ternary representations, that have demonstrated reducing a LLM (large language models as in ChatGPT, Grok, DeepSeek) from a 700-watt NVIDIA-GPU workload, down into 13 watt commodity FPGA chip (127). OpenGPU (https://www.OpenGPU.com) seeks to invest in solutions for customer-defined problems and projects that fill funding gaps between NSF, DOE and other agencies, with focus on commercialization and engineering rigor.

## Federal Funding Opportunities

The BRAIN Initiative is disease-agnostic and focuses on developing new tools, theories and methods, and collecting large-scale datasets, which it is interested in leveraging in convergent approaches and technologies at the intersection of AI and neuroscience; see the NIH Highlighted Topic that encourages relevant proposal submissions to parent announcements at https://grants.nih.gov/funding/find-a-fit-for-your-research/highlighted-topics/18. Additionally, NeuroAI-related applications may be submitted to BRAIN Initiative: Theories, Models and Methods for Analysis of Complex Data from the Brain (R01 Clinical Trial Not Allowed); see https://grants.nih.gov/grants/guide/rfa-files/RFA-DA-27-004.html.

NINDS supports research using big data approaches and developing devices to provide real-time, real-life data, with funding opportunities including Translational Neural Devices (R61/R33), Bioengineering Research Grants (BRG), and Collaborative Opportunities for Multidisciplinary, Bold, and Innovative Neuroscience (COMBINE). For more, visit https://www.grants.gov/ and search for NIH opportunities via the *Keyword(s)* or *Opportunity Number* fields.

NIBIB sits between fundamental science/engineering at NSF and clinical translation at other NIH institutes, supporting broad platform technology and biomedical applications across imaging, bionics, prosthetics, wearables, and AI/ML. Funding could be secured through many mechanisms including parent R01, R21, R03, and P41. For more, visit https://www.nibib.nih.gov/funding.

The Brain Behavior Quantification and Synchronization (BBQS) program integrates dense recordings of neural signals, behavior, and environment, with particular interest in energy-efficient sensor technologies for minimally invasive monitoring. For more, visit https://brain-bbqs.org/.

NIDA's data science and informatics program supports data sharing, infrastructure, standards development, and integration tools for BRAIN Initiative research, with multiple web-accessible data archives already established. For more, visit https://nida.nih.gov/about-nida/organization/divisions/division-neuroscience-behavior-dnb/data-science-research

The NIH Common Fund's Precision Medicine with AI: Integrating Imaging with Multimodal Data (PRIMED-AI) program will combine clinical imaging with other types of health data ("multimodal data") to develop innovative artificial intelligence (AI)-powered clinical decision support (CDS) tools to enable new personalized medicine strategies. For more, visit https://commonfund.nih.gov/primed-ai.

NSF ENG Directorate has invested ~$585 million in 875 brain-inspired research awards and ~$37 million in 65 neuromorphic projects, with the Division of Electrical, Communications and Cyber Systems managing ~$107 million in 225 brain-inspired research awards and ~$19 million in 50 neuromorphic projects (50 awards). The momentum in this space as reflected by recent investments made by NSF EFRI programs including BRAID which invested ~28 million in 16 awards ((128), Table 1) and Biocomputing through Engineering Organoid Intelligence (BeginOI) which focused on understanding fundamental biological principles and the use of biological constructs with hardware interfaces for computing rather than exact biological mimicry to address unsustainable energy demands of current computing approaches. For more, visit https://www.nsf.gov/funding/opportunities/emerging-frontiers-research-innovation-efri-biocomputing/13708/nsf24-508.

NSF CISE Directorate's Foundation of Emerging Technologies (FET) program focuses on going beyond von Neumann computing through biocomputing, quantum computing, and neuromorphic computing approaches. For more, visit https://www.nsf.gov/funding/opportunities/fet-ccf-foundations-emerging-technologies.

NSF's SBIR program provides up to $2 million to seed initiatives that translate scientific breakthroughs into products or services, funding high-risk, high-impact technology development. For more, visit https://seedfund.nsf.gov/.

Army Research Laboratory seeks basic foundational science opportunities to enhance soldier/human systems performance, with programs in Neurophysiology of Cognition and Bionic Electronics supporting high-risk projects without preliminary data requirements. The funding includes efforts to understand brain computations for groundbreaking development of theories, algorithms, and hardware, as well as efforts to develop novel hybrid computing systems to harness the advantages of both living and non-living networks. For more, visit https://www.arl.army.mil/collaborate-with-us/opportunity/arl-baa/.

DARPA pursues neuro-inspired computing through multiple offices including I2O (AI and cybersecurity), MTO (microsystems including scalable analog neural networks), and DSO (Defense Sciences Office with broad solicitations). For more, visit https://www.darpa.mil/work-with-us/opportunities.

A new NSF-NIH biphasic award mechanism called "BRING SynBio" allows projects to start with NSF funding for two years then transition to NIH for two years, currently focused on synthetic biology but potentially expanding to other areas. For more, visit https://www.nsf.gov/funding/opportunities/bring-synbio-biomedical-research-initiative-next-gen-biotechnologies/nsf24-603/solicitation.

Existing NSF-NIH collaborations including the Collaborative Research in Computational Neuroscience (CRCNS) program, and the Smart Health and Biomedical Research in the Era of Artificial Intelligence and Advanced Data Science (SCH) program are also in scope for neuromorphic approaches. For more on CRCNS: https://www.nsf.gov/funding/opportunities/crcns-collaborative-research-computational-neuroscience. For more on SCH: https://www.nsf.gov/funding/opportunities/sch-smart-health-biomedical-research-era-artificial-intelligence.

## Day 2: Keynote – Learning from Skin: from Materials, Sensing Functions to Neuromorphic Engineering

Zhenan Bao, K.K. Lee Professor of Chemical Engineering of Stanford University, presented her work on skin-inspired materials and electronic systems that process signals at the sensor interface. This mimics how skin mechanoreceptors respond to external stimuli through frequency-encoded spike train signals. Bao's team created breakthrough materials with nano-confined structures that force polymers to adapt planar configurations, significantly enhancing charge transport mobility compared to traditional semiconductor materials. The research has produced a collection of skin-like electronic materials including stretchable semiconductors, biodegradable conductors, self-healing polymers, and transparent electrodes with orders of magnitude lower impedance than metal electrodes when in contact with tissue (60).

Bao's group has developed integrated circuits with over 1,000 transistors on stretchable substrates operating at megahertz frequencies, enabling signal processing directly at the sensing front. This led to the creation of pressure sensors with high spatial resolution (less than 500 microns) that exceed human tactile resolution, demonstrating the potential to develop electronic skins with capabilities beyond biological limitations including accuracy improvements in Braille reading (60). Her research demonstrated neuromorphic signal processing by using ring oscillators to generate spike train signals with pressure information encoded as frequency and synaptic transistors that combine multiple sensor inputs (129). For implantable applications, the team developed "NeuroString" - one-dimensional flexible probes with 150-300 sensors in a narrow (150-250 μm) diameter structure that can be implanted in the brain and gut for simultaneous neurotransmitter measurements in rodents (130).

She then introduced the latest version of her electronic skin, which she coined "neuromorphic skin" where the frequency encoded signal, the sensor, and circuits are all monolithically integrated into a sensing sheet where pressure or temperature information could be encoded. This system is devoid of any rigid electronic components, can emulate the sensory feedback functions of biological skin, including multimodal reception, nerve-like pulse-train signal conditioning, and closed-loop actuation (13). She reminded the audience that most of the materials that she develops are relatively new with uncertain long-term biocompatibility. Bao concluded by identifying key challenges including the need for long-term biocompatibility testing, challenges in prototyping nonconventional materials, and determining which applications would benefit most from the hybrid or neuromorphic design approach. As a multidisciplinary community, we need to understand the strengths and the limitations for different design approaches.

## Day 2 Keynote – Q&A Highlights

When questioned about what constitutes "neuromorphic," Bao clarified that her approach mimics spike train communication rather than precisely replicating neural action potential profiles. Focusing on a simplistic understanding of neurons and synapses results in simpler circuit designs. When addressing power efficiency, Bao noted that the designed transistors have a lower charge carrier mobility than crystalline silicon. However, the transistors also conduct less current with minimal heating during operation. Potential applications were discussed, including enhanced tactile sensing for humans as well as robots, prosthetics with cybernetic ports, skin-to-skin communication modalities, and flexible sensors for the gut.

## Day 2: Session 3 Materials for Neuromorphics (Devices)

Session 3 was moderated by Dr. Shantanu Chakrabartty of Washington University St. Louis and Dr. Duygu Kuzum of the University of California San Diego (UCSD) and covered multimodal neural interfaces and neuromorphic co-design, Brainchip's neuromorphic device, and novel materials for biomedical applications and neuromorphic computing.

The first speaker was Dr. Duygu Kuzum. Her talk, *Innovating Beyond Electrophysiology: Multimodal Neural Interfaces and Neuromorphic Co-Design*, addressed limitations in current neural recording technologies and explained the value of neuromorphic co-design. While electrophysiological recording technologies have made steady progression, recording capacity is still a limitation, with even advanced Neuropixel probes only capturing about 1,000 neurons simultaneously. Optical imaging techniques, in comparison, have doubled in recording capacity approximately every 1.5 years and can now image tens of thousands of neurons across 3D brain volumes (131). Current neural interface challenges include tissue damage during implantation, limited spatial resolution of recordings, and bulky head-stage equipment for optical imaging.

Kuzum has developed transparent graphene-based microelectrodes that allow for multimodal experiments by integrating optical imaging, optogenetic stimulation, and electrophysiological recordings without crosstalk (61). These transparent electrodes simultaneously monitor neural activity at multiple spatial scales, including recording local field potentials from the cortical surface while monitoring calcium spikes from deeper layers (132). For real-time neural signal processing, Kuzum has developed neuromorphic brain machine interfaces based on resistive random access memory (RRAM),--also called memristor crossbar arrays--to perform efficient, in-memory spike sorting (28). She has demonstrated that RRAM crossbars could process recordings from 100 channels within 4.8 microseconds while consuming 30-50 times less energy than FPGA approaches (62).

Kuzum has also led research on neuromorphic co-design to extrapolate additional spatial information from neural recordings with neuro-inspired approaches. Neural recordings from the surface of the brain include spatially integrated data of the neural activity further below the electrode. Using neuromorphic computational models with recurrent neural networks, Kuzum successfully decoded individual cellular calcium activity in deeper brain layers using only surface potentials, expanding spatial reach (61).

Kuzum concluded that neuromorphics can help neuroscience and neuromedicine in two major ways: [1] neuromorphic computational models can integrate information from multiple modalities to create accurate, low dimensional neural representations to improve neural prosthetics or BMI performance; [2] neuromorphic hardware can transform neural recording methodologies by enabling on-chip, real-time signal processing with learning capabilities that can shift the processing location of the data and increase longevity and stability of neural interfaces.

The second speaker was Dr. Tony Lewis, CTO of BrainChip. His talk, *Flow Machines and Neuromorphic Devices*, presented BrainChip's commercial neuromorphic technology. The Akida 2.x chip was featured, which is a digital, event-based, near-memory compute system. The Akida architecture has a distributed array of processors connected in a mesh network. Each processor contains four nodes and each node contains a Temporal Event-based Neural Networks (TENNs) processing unit. Event-based convolution dramatically reduces computational requirements by only processing non-zero values. For example, event-based convolution can reduce the multiply-accumulate operations (MACs) from 225 to 27, while computing the same result. Lewis introduced the concept of "flow machines" to distinguish future edge AI (which includes neuromorphics and TENNS) from conventional AI systems that leverage state memory. BrainChip's TENNs model is a state space model that leverages event-based computation and can replace transformer tasks, such as language models, spatiotemporal data, and time series data, while dramatically reducing power consumption. The approach uses Chebyshev polynomials to compute convolutions in kernel mode or recursive mode, allowing for training of extremely deep networks (100+ layers). BrainChip's TENNs have many uses including audio denoising, audio keyword spotting, generative AI large-language models, and industrial artificial intelligence of things. The audio keyword spotting operates at 0.25 milliwatts with 97% accuracy, and the audio denoising system can remove background noise from speech samples while maintaining intelligibility, potentially leading to "super intelligibility" applications. BrainChip's TENNs achieved better perplexity scores (ability of a large-language model to predict the next word) than Mamba and Llama 3.2 with a fraction of the data at half the model size. Lewis emphasized that successful neuromorphic implementation requires both clever hardware and software that can take advantage of that hardware to create sparse networks.

The third speaker was Dr. Deji Akinwande of the University of Texas at Austin. His talk, *Towards Health Monitoring & Computing using Advanced Materials*, focused on fundamental materials, particularly two-dimensional (2D) materials like graphene, which are atomically thin, flexible, stretchable, and potentially biocompatible for biomedical applications and neuromorphic computing. These 2D materials are less than a nanometer thick and offer multifunctional properties as electrodes, sensors, amplifiers, and transistors. Akinwande's lab explored these materials for both "organ-tronics" (*in vivo* sensing) and "skin-tronics" (direct skin interfacing) by building on prior work demonstrating graphene's effectiveness for implantable sensing and electrophysiological monitoring (63). Graphene's transparency enables combined high spatial resolution optical imaging with high temporal resolution electrical recording, and can also be used for treatment applications such as a transparent graphene pacemaker (64). Akinwande has developed ultra-thin graphene electronic tattoos that are barely detectable on skin and can be used as a wrist-worn bioimpedance sensor for blood pressure monitoring. In addition, he achieved clinical-grade accuracy of blood pressure monitoring through an AdaBoost regression model (65). With respect to neuromorphic materials, Akinwande has created "atom-resistors" (memristors in atomically thin materials) that demonstrated both binary and analog resistance switching (66). These atom-resistors are suitable for neural networks, with classification accuracy approaching theoretical limits in an MNIST digit recognition task. He also developed flexible synaptic transistors using graphene with nafion (a fluoropolymer ion conductor) as the dielectric layer, enabling potentiation and depression functionality with extremely low energy dissipation compared to other materials (133). The most intimate interface between synthetic and biological materials that Akinwande's team has created is a "skin-gated transistor for neuromorphic tattoos". Here, the human skin directly serves as the gate dielectric layer of the transistor, allowing synaptic weight changes directly through skin interface (unpublished work). Their future goal is to move from offline processing to local sensing and computing to enable direct monitoring of physiological indicators, like sodium or calcium ions, with minimal device requirements.

## Day 2 Session 3: Panel Discussion Highlights

Gina Adam of George Washington University, Jennifer Hasler of Georgia Tech, Gert Cauwenberghs of UCSD, and Ulkuhan Guler of Worcester Polytechnic Institute joined the panel along with the speakers and session co-chairs.

Shantanu Chakrabarty highlighted the importance of embodiment and control in hybrid neural interfaces. Cross mapping between biological and synthetic neurons is missing. He pondered if biological neurons would treat silicon neurons as their own. Jennifer Hasler argued that neuromorphic applications or wearable, implantable system could very well be done entirely within standard complementary metal-oxide semiconductor (CMOS) (7). Gert Cauwenberghs emphasized returning to the physical foundations that Carver Mead established when coining "neuromorphic engineering," noting that despite different substrates (biology vs. silicon), similar physical principles govern both systems (4,5). Both biological ion movement through channels and electron/hole movement in transistors are governed by the same foundational physics - Boltzmann statistics and thermodynamics. These principles extend to higher levels, including network-level behaviors and energy-based optimization dynamics. The field benefits from complementary top-down and bottom-up approaches that create synergies between computational systems neuroscience analysis and neuromorphic systems engineering synthesis. This approach enables scaling from carriers up to whole brain-level interfaces with isomorphism at every level and has led to more efficient neural interface circuits inspired by neuromorphic computing principles (134). Ulkuhan Guler described a biomorphic transcutaneous sensing system that her lab prototyped to detect oxygen and carbon dioxide molecules in a patch or watch formfactor. Gina Adam expressed the need for: 1) datasets to co-design materials, devices, and algorithms that can compute locally. This requires a closed loop between sensing, computing, and actuation; 2) metrics, including energy efficiency, compactness, noise, and robustness of the hardware and software; 3) interdisciplinary collaboration and education.

When asked about new neuromorphic opportunities in using new materials, panelists discussed the importance of biocompatibility and CMOS integration for new materials. Deji Akinwande suggested that the human body should be integrated as an active component of the device rather than just viewing the human body as a passive target to probe, treat, and/or study.

When discussing whether neuromorphic is the only solution to power and bandwidth challenges, Shantanu Chakrabarty emphasized identifying specific applications where neuromorphic provides clear advantages (e.g., latency to achieve real-time closed-loop systems, extremely efficient and resilient circuits).

When asked about challenges in manufacturing neuromorphic chips for biomedicine, Tony Lewis explained that he targets CMOS because it is widely available. Other manufacturing challenges include split foundry issues, tool chain limitations, and the high costs associated with chip fabrication, with estimates ranging from $2-30 million just to start a chip (while others argue that a few thousand dollars on a multi-project wafer run is sufficient at large volume production levels).

When asked about benefits of on-chip learning, Duygu Kuzum explained that neural signals drift over time for a variety of biological and non-biological reasons. On-chip learning could be a fast and efficient way to correct for signal drift; this would benefit long term experiments that track the same type of neural dynamics over time to study behavior and cognition in both animal and humans. Ulkuhan Guler added that interpersonal and intrapersonal variations could also be addressed by on-chip learning. Gert Cauwenberghs offered that from a circuit perspective, one additional advantage of adaptation is to circumvent imperfections.

When asked about challenges in clinical translation, panelists explained that a key challenge is that long investment horizons do not align with typical business expectations, as well as regulatory pathways, privacy concerns, and security considerations.

## Day 2 Session 4: Medical Imaging, Wearables and Analysis

Session 4, moderated by Jennifer Blain Christen of Arizona State University and Ralph Etienne-Cummings of Johns Hopkins, covered wearables, biomedical interfaces, analysis, and computation using neuromorphic devices.

The first speaker was Sihong Wang of the University of Chicago. His talk, *Bio-electronic interfaces: from high fidelity biosensing to integrated neuromorphic edge computing,* presented new bio-electronic interfaces that enabled precision medicine through large-scale physiological data monitoring and neuromorphic computing. He asserted that current commercial wearable and implantable devices face several challenges that limit data collection capabilities. These include limited function and lack of access to important physiological biomarkers in wearable devices and reduced lifespan of implantable devices (typically < 2 years in the case of brain electrodes). Academic research is increasingly utilizing AI to analyze wearable and implantable data, but the raw sensor data is typically processed through cloud computing, which increases latency, consumes more power, and risks privacy breaches. He posed two grand challenges for the field as follows: 1) how to develop conformable, elastic, low profile sensors for collecting high-fidelity, multi-modal, and long-term stable data at skin/tissue surfaces? And 2) how to incorporate neuromorphic edge computing functions into wearable/implantable systems?

To address grand challenge #1, Wang has developed a soft, stretchable, immuno-compatible, bioadhesive, bioelectronic interface to improve the signal-to-noise ratio of the sensor. The properties of the interface prevent signal decay by conforming to the tissue and preventing fibrosis for long-term implants. The hydrogel interface is comprised of an immune-compatible, semiconducting polymer, which resulted in a suppressed foreign body response (FBR) when implanted subcutaneously for 4 weeks (135); when selenophene is incorporated suppression of the FBR reached 68% (82). An organic electrochemical transistor (OECT) was designed using the hydrogel semiconductor to record both electrical and chemical signals. This bioelectronic interface can directly attach to tissue surfaces, like a beating rat heart, and maintain stable signal quality through strong mechanical interfacial adhesion (136).

Regarding grand challenge #2, Wang demonstrated a fully stretchable, neuromorphic computing 3x3 array with integrated memory and computing functions directly on the hardware using OECTs. The neuromorphic device has high classification accuracy for abnormal ECG signals and maintains that accuracy during increasing mechanical strain (30). He then featured Gina Adam's work on neuromorphic detection of propagation wavefront in cardiac ventricular fibrillation to trigger precise ablation (137) as an inspiration to his recent research on large-scale, stretchable neuromorphic circuits for on-body edge processing of sensory data for precise treatment delivery (67).

He concluded his talk by posing four additional questions for the field to resolve, as follows:

(1) Which part of machine learning (ML) / neuromorphic data processing functions should be physically integrated with sensory devices (i.e. edge vs cloud computing)?
(2) How can we better physically integrate and functionally merge sensing functions with neuromorphic computing?
(3) Which type of ML algorithms and neuromorphic device types could be the most suitable for integrating with sensory devices to provide the lowest complexity and power consumption?
(4) What should be the design of peripheral circuits that also have tissue / skin-like properties?

The second speaker was Erika Schmitt of Cambrya. Her talk, *Novel neuromorphic algorithms for adaptive learning at the edge*, focused on the auditory blind source separation problem using synchrony loop propagation through bidirectional signal exchange across graded spikes in density networks (71). Cambrya's network architecture was inspired by the auditory system with tonotopic organization and increasingly complex cognitive functions. As a result, this network architecture can separate multiple audio signals from crowded, noisy environments with no prior training data, a learning time of 0.5 seconds, and a model size of only 12,000 parameters. While over 1.5 million people globally are affected by some form of hearing loss, few people with mild to moderate hearing loss use assistive devices. Cambrya's technology can provide an alternative to address hearing loss by allowing users to select and amplify specific voices in crowded environments instantaneously. Learning is happening directly on the device with no cloud processing required. Her team is pursuing market entry through traditional hardware with a flexible SDK that can integrate with existing hearing devices. Long term, Cambrya would like to implement their architecture on neuromorphic hardware. Finally, Schmitt noted that the neuromorphic nature of their algorithm is no longer divulged to potential investors because the term "neuromorphic" makes the technology appear more risky. Therefore, there is a need in this current funding ecosystem to generate excitement around neuromorphics and their near-term, commercial capabilities.

The third speaker was J. Brad Aimone of Sandia National Laboratories. His talk, *From new neurons to new chips: a path to a new way of thinking about the brain?*, discussed how neuromorphic computing can help us understand neuroscience better by moving beyond traditional sequential processing models to more brain-like computational frameworks. He observed that current theoretical neuroscience frameworks as overly simplistic, often applying the same fundamental principles to different brain regions despite their diverse architectures and functions. Aimone's team has explored spatial-temporal processing approaches that trace causal chains through spiking networks, leveraging structural relationships from connectomes rather than relying solely on sequential time-based analyses. They have found that neuromorphic systems excelled at sampling problems (i.e., Monte Carlo methods (68)) and physically mapping spatial interactions (i.e., finite element approaches (69)), which may reflect fundamental computational principles of the brain. Their neuromorphic approach to solving partial differential equations showed dynamics remarkably similar to biological neural recordings, despite drastic differences in architectures. Aimone concluded that "what makes neuromorphic hardware special are the things that make brains special." Both utilize heterogeneity, parallel processing, and spatial temporal processing. Thus, these attributes deserve more focus when creating theoretical frameworks for understanding how the brain works.

## Day 2 Session 4: Panel Discussion Highlights

Andreas Andreou and Robert Stevens of Johns Hopkins University, Pamela Abshire of the University of Maryland, College Park, and Amparo Güemes of the University of Cambridge in the United Kingdom joined the panel along with the speakers and session co-chairs. Andreas Andreou highlighted the importance of constructing algorithms that exploit temporal processing or time, and the need to invest in research and design methodologies, like co-design algorithms with synchronous circuits or event-based circuits. However, design automation tools are not able to support automated workflows from algorithm-to-chip design. He asserted that neural chiplets are the way to go, as they are modular, scalable, and reusable. Neural chiplets can be integrated into heterogeneous devices of various complexities with different materials and structures. He concluded with a newly released funding opportunity called [National Advanced Packaging Manufacturing Program](#) (94) and suggested that the NIH pays attention to this funding model.

The panel discussed challenges in bringing neuromorphic technologies to healthcare, including regulatory barriers, integration with existing clinical workflows, barrier with multiple disciplines needed to be successful (neuroscience, materials, data science, engineering), and clear demonstrations of the benefits of neuromorphics. Power requirements were debated, with many participants arguing that neuromorphic approaches (low power, low heat dissipation) are essential for implantable devices, while a few suggested the power advantages might be oversold for many medical applications. In discussing flexible bioelectronics, panelists highlighted the potential for neuromorphic approaches to enable better sensor fusion across modalities (molecular, pressure) and different time-scales. Panelists strongly identified a need for new design tools, fabrication platforms, systems integration beyond chip design, and standards for neuromorphic devices to accelerate development and reduce time spent on custom implementations. The term "neuromorphic" was noted as a potential hinderance towards technology adoption, and suggestions were made to focus instead on demonstrating the "magical" capabilities that cannot be achieved with conventional approaches.

## Day 2: Moderated Discussion

*Opening Discussion on Benchmarks and Data*
Workshop participants identified the need to develop specific, standardized datasets rather than focusing on benchmarks, which some participants found to be restrictive. Brad Aimone argued that benchmarking, despite its popularity in AI development, would be counterproductive for the neuromorphic community by forcing developers to use outdated paradigms while the field of neuroscience moves forward. Andreas Andreou offered historical context from computer vision, noting that benchmarks initially helped standardize evaluation. However, innovation was eventually constrained since researchers optimized specifically for benchmark datasets rather than solving real-world problems. The discussion highlighted the precarious balance between acknowledging the value of standardized datasets while avoiding restrictive benchmarks.

Several participants emphasized the importance of simultaneous, multimodal recordings with different sensor types (e.g. EEG, EOG, EMG, ECoG) capturing the same events. This would allow for correlation across multiple modalities and facilitate algorithm development without constraining innovation. Andreas Andreou noted that researchers such as Parisa Rashidi are collecting data from ICUs potentially with event-based cameras that could be of great value to the neuromorphic community. It is critical to look outside the neuromorphic world to the broader community for datasets. Sydney Cash noted that substantial intracranial recording datasets already exist through BRAIN data repositories like DABI and DANDI. Also, these datasets have been previously leveraged in algorithm development competitions like Kaggle to solve biomedical problems such as seizure prediction. Multiple participants suggested that future algorithmic or data challenges using these multimodal datasets could feature neuromorphic algorithms or hardware and the specific advantage over conventional approaches.

*Clinical Need: Sensor Technology, Device Specifications, and Data Collection*
Participants discussed the need for better wearable sensors that move beyond the limitations of the clinic and collect data while subjects are mobile and engaged in natural behaviors. Cornelia Fermuller highlighted event-based vision sensors in the neuromorphic community (e.g., https://inivation.com/) as a successful technology with practical applications in low-power eye tracking and action recognition. Dana Schloesser mentioned the BRAIN BBQS (Brain Behavior Quantification and Synchronization) program, which aims to capture synchronized neural and behavioral data in naturalistic environments across both clinical and non-clinical populations. It was noted that companies like Tucker Davis Technologies provide multimodal recording systems including electrophysiology, fiber photometry, auditory, and behavior on preclinical animals. This discussion highlighted important device specifications including biocompatibility, stretchability, adhesiveness, power consumption, latency requirements, and the ability to either dissolve or remain stable in the body (e.g., heart versus stomach) depending on the application.

William Stacey provided specific technical requirements for neural recordings in ambulatory and inpatient acute setting that would benefit neurology, psychiatry, and neuromodulation applications. These capabilities include customizable electrode placement, stimulation parameters, and sampling rates (0.5 - 4000 Hz), as well as data collection on continuous recordings.

Several clinicians emphasized the need for technologies that monitor multiple biomarkers. This includes chemical signals, such as potassium, oxygen, sodium, and various neurotransmitters (up to 27). Robert Stevens specifically identified the immune system as a critical but under-surveilled biological system that lacks continuous, real-time monitoring. Tim Denison noted that companies such as SetPoint Medical and Galvani have ongoing clinical trials in the immunomodulation space, cautioning against reinventing the wheel. Participants discussed the need for multimodal recording devices that can simultaneously measure electrical activity alongside chemical signals to provide a more complete picture of neural function.

*Regulatory Considerations and Barriers*
Participants called for better education on regulatory processes for adaptive systems. Tim Denison noted the existence of regulatory guidance documents for physiologic control systems (72), document # 60601-1-10 (73), and AI-based medical devices (126) that outline important regulatory considerations for developers. Participants identified manufacturing challenges specific to neuromorphic devices, including testing methods for novel components like memristor arrays and ensuring reliability in the harsh biological environment where temperature fluctuations and the foreign body response can affect performance.

Pam Abshire raised concerns about regulating adaptive systems, noting the difficulty in quantifying the benefits of adaptation and establishing appropriate regulatory frameworks for systems that learn and change over time. Shantanu Chakrabarty discussed the importance of fail-safe mechanisms in adaptive systems, similar to autonomous vehicles with different levels of autonomy, to ensure devices can revert to baseline functioning if problems arise. Other participants suggested that neuromorphic systems are resilient to hacking, when compared to today's electronics, owing to the stochastic nature of the brain. This led to a discussion on the stability plasticity dilemma and how would the FDA regulate an unstable system. Tim Denison noted that when NeuroPace RNS was first launched, similar concerns arose, and dose limits were employed as guardrails to limit risk to the user should the device go awry.

*Infrastructure Needs*
Multiple participants emphasized the need for better infrastructure to support neuromorphic engineering, including fabrication access, design tools, protocol sharing, and standardization. There was discussion about limited access to fabrication facilities and a call to establish programs similar to the historical MOSIS initiative that provided free chip manufacturing for academic researchers (77). Andreas Andreou suggested expanding beyond silicon chips to incorporate new materials and structural complexity (e.g., bendable bio-electronics), with a supporting infrastructure equivalent to MOSIS for these novel approaches. Sunny Bains reiterated Jennifer Blain Christen's suggestion for protocol sharing to eliminate the need to start from scratch, especially when developing novel biohybrid neuromorphic materials. Ulkuhan Guler noted the importance of chip design to ensure US competitiveness in the global market. Cornelia Fermuller identified the lack of design tools for the neuromorphic community as a critical gap. Development would accelerate if tools for designing asynchronous circuits and transferring spiking neural networks between hardware platforms were available. Grace Hwang pointed out the significant gap between the design tool ecosystems available for traditional digital signal processing versus neuromorphic computing, emphasizing that substantial work remains to make neuromorphic approaches more accessible and scalable.

# Appendix 2: Panelists' Position Statements

*Topic 1 - Epilepsy & Cortical Disorders (Neuromodulation)*
Viktor Jirsa: In the past twenty years, we have made significant progress in creating digital models of an individual's brain, so called virtual brain twins. By combining brain imaging data with mathematical models, we can predict outcomes more accurately than using each method separately. Our approach has helped us understand normal brain states, their operation and conditions like healthy aging, dementia and epilepsy. Using a combination of computational modeling and dynamical systems analysis we provide a mechanistic description of the formation of resting state manifold via the network connectivity. We demonstrate that the symmetry breaking by the connectivity creates a characteristic flow on the manifold, which produces the major data features across scales and imaging modalities. These include spontaneous high amplitude co-activations, neuronal cascades, spectral cortical gradients, multistability, and characteristic functional connectivity dynamics. When aggregated across cortical hierarchies, these match the profiles from empirical data and explain features of the brain's microstate organization. Examples for clinical translation are taken from drug resistant epilepsy and mental disorders. The digital brain twin augments the value of empirical data by completing missing data, allowing clinical hypothesis testing and optimizing treatment strategies for the individual patient. Virtual Brain Twins are part of the European infrastructure called EBRAINS, which supports researchers worldwide in digital neuroscience.

Kendall Lee: Modern neural engineering technologies now allow for control of neurotransmitter levels by DBS. Artificial Intelligence will allow for automated neural control algorithms in the future. Further complex electro-analytical methodologies such as MCSWV, will also be available for neuropsychiatric therapies.

Chris Rozell: Currently, commercial neurotechnologies are designed with very minimal processing capability, geared almost entirely toward very specific notions of what onboard processing is necessary for closed-loop stimulation to treat specific disorders such as epilepsy and PD. There are presumably many constraints driving this narrow product design, including power consumption, form factor limitations, and the time necessary for new engineering designs that meet regulatory standards. Future innovations would benefit from an expanded set of capabilities in a package that improves the user experience in ways that will increase accessibility (e.g., smaller form factors, longer battery life, etc.).

Sridevi Sarma: During routine physical examinations, various physiological parameters such as age, weight, height, and blood pressure are measured to gauge overall health status. Among these, Body Mass Index serves as a widely utilized tool to stratify individuals into different weight categories, aiding in the identification of potential health risks. However, a pertinent inquiry arises regarding the identification of indicators for brain health and whether a quantitative assessment of brain health can be derived from imaging data. In this proposed study, we develop the "Brain Entropy Index (BEI)," a novel quantitative measure of brain health, ranging between 0 to 1, where a high BEI (e.g., > .75) would indicate a healthy brain and a low BEI would indicate a risk for a neurological disorder. The BEI is not derived by training machine learning models on large data sets, which is the typical approach used today to diagnose neurological disorders from imaging data. In contrast, the BEI is derived from a unifying theory that combines principles from thermodynamics, dynamical systems and matrix theory. This unifying theory (i) identifies properties that a healthy human brain at rest must have, and then (ii) derives metrics that characterize whether these properties exist in a patient using imaging data. The idea is to capture resting-state EEG for 10-15 minutes from patients visiting their primary care physicians for annual exams and compute the BEI in minutes after the EEG scan is captured. Furthermore, we illustrate how deviations in BEI can effectively classify various neurological disorders, including epilepsy, Alzheimer's disease, and frontotemporal dementia. The BEI also facilitates the assessment of treatment efficacy, with successful treatments restoring the BEI to a healthy range. Finally, we showcase how linear time-varying models can be leveraged to simulate virtual surgeries for epilepsy based on neurosurgeon's plans, with surgical success predicted by measuring the BEI pre- and post-virtual surgery.

William Stacey: High Frequency Oscillations (HFOs) have been investigated for over 20 years as a potential biomarker of epileptic tissue. However, HFOs are also present in normal brain. There has been great interest in distinguishing normal from pathological HFOs. Early studies on small datasets suggested that HFOs could be classified by peak frequency, a sentiment that persists to this day. However, many later rigorous studies have refuted this, and there is no true gold standard for comparison. We collected a large dataset with millions of HFOs and used several strategies to find pathological features. We find consistent features in many patients, but the large heterogeneity between patients makes classification very difficult. Larger datasets and better controls (which are extremely difficult to acquire) are needed for better classification.

Sydney Cash: Neural interfaces, neuromodulatory systems and brain-computer interfaces, have made significant strides over the past decade, evidenced by ongoing clinical trials and increasing clinical applications. However, substantial challenges remain in harnessing their full potential for clinical use. On the engineering side, challenges such as energy efficiency, packaging, data management, and security are identified and the focus of research in multiple groups. Conversely, our understanding of human brain physiology and its interface with these technologies often lags behind our engineering capabilities, which translates to a poor grasp of the fundamental constraints underlying neuromorphic approaches. For instance, in well-studied conditions like epilepsy, we still grapple with critical questions: Where do seizures originate? How do they start? What is the optimal scale for interacting with and mapping seizure activity? What areas should we target for modulation? Furthermore, we remain uncertain about the short- and long-term cellular responses to various forms of neuromodulation. To fully leverage the potential of brain-computer interfaces and neuromodulatory technologies, continued investment and interdisciplinary dialogue are essential. This cannot be achieved without integrating engineering advancements with biological understanding in a collaborative feedback loop.

Nathan Crone: Recent clinical trials of implantable brain-computer interfaces have shown that both single unit activity and electrocorticographic signals can be decoded into text and speech to restore communication in people with varying degrees of Locked In Syndrome. These achievements have relied heavily on advances in machine learning that were originally inspired by neuromorphic principles. However, current decoding models require vastly greater computing power than is practical for fully implantable, wireless BMI systems.  Moreover, the limitations of currently available neural interfaces have highlighted the need for finer and more comprehensive sampling of neural elements, further inflating demands on data transmission and computation. Neuromorphic principles could potentially be used in the design of neuroprosthetic devices that meet these growing needs.  Furthermore, these principles may be particularly helpful when moving beyond decoding the neurophysiological signatures of brain health and disease, to restoring healthy brain function through effective neural stimulation.

Giacomo Valle: The needs for the neuromorphic technologies in neuroprosthetics relate to the exploitation of biologically inspired process to interface, communicate and interact with the nervous system at multiple levels. Neuromorphic tech could potentially allow for a better information transfer between the artificial and biological systems. Moreover, the high level of portability and embeddability could boost the transability of these technologies into clinical applications (24). The challenges are related to the exploitation of the sophisticated neuromorphic framework given the limits of the current available interfacing technology. Moreover, the neuromorphic technology is based on the emulation of the biological counterparts. Notably, important natural processes related to the target neural structures have not been defined or not exhaustively investigated yet, making the validation quite challenging. The contribution of neuromorphic tech has shown interesting applications in artificial sensing (vision and touch) and computing. The flexibility of the technology, its embeddability, could favor its application in everyday life settings (e.g. prosthetics). The benefit to exploit a computation based on neural emulation would allow for a better human-machine interfacing. Human neural processing has been shaped by evolution over thousands of years, and it is flexible and efficient. Its emulation through neuromorphic tech opens important applications in engineering and medicine (neuromodulation, neuroprosthetics, etc.). The design of a future generation of bio-inspired neuroprosthetic devices allowing the artificial conveyance of more complex information and therapies with neurostimulation could drastically improve the field. The neuromorphic tech could potentially increase, not only the efficacy but also the acceptance of more bio-inspired neuroprosthetic devices. The use of neurostimulation is also critical to brain–computer interfacing, as well as bioelectronic medicine, in which the electrical stimulation targets the central, or the autonomic, nervous systems, respectively. As the healthcare sector increasingly adopts AI and advanced computational methods, neuromorphic technologies are likely to play a pivotal role, making them a compelling area for investment. Investing in neuromorphic technologies in medicine presents several promising opportunities: neuroprosthetics, diagnostics, mental health applications, wearable devices, rehabilitation technologies, and AI in medicine.

*Topic 2- Human-machine interfaces (Prosthetics)*

Xing Chen: Needs: Clinical solutions that are safe, long-lived and effective, providing assistance in daily tasks, object recognition, navigation, and independence.  Challenges: Technological challenges in interfacing with the brain with sufficient resolution and specificity, allowing neuromodulation without inducing seizures and aberrant activity, building devices that are easy to implant across clinical sites and individual patients.  Contributions: Development of biocompatible materials, scientific understanding of neural encoding and decoding to improve specificity and efficacy of neuromodulation techniques.  Impact: Providing hope and inspiration for blind and visually impaired people, and laying the foundation for breakthrough devices.  Investment: Harnessing solutions from the lab and bringing them to patients. Recruitment of patients and other stakeholders to shape and guide the field from an early stage. Raising awareness amongst the general public of both the promise and pitfalls of this technology.

Helen Huang: 1. The design principles for neural-machine interfaces (NMIs) should be anchored in established human motor control and learning theories. 2. The design of an efferent neural decoder should be guided by human neuromechanics. 3. The design of an afferent neural encoder should aim to enhance sensorimotor integration 4. Neuromorphic computing and artificial intelligence are powerful tools for addressing NMI adaptations to individual users and for evolving over time, with low power consumption and distributed processing.

Ranu Jung: Neurotechnology holds the promise of enhancing our understanding and offering innovative treatment of both physical and mental health conditions. However, transitioning these advancements from the laboratory to clinical and home settings to transform healthcare remains a significant challenge. This talk discusses the use of a convergence approach for pioneering advancements in neurotechnology to revolutionize healthcare. The journey from concept to development and deployment of a Neural-Enabled Prosthetic Hand System in an FDA approved first-in-human trial will be used to illustrate the extensive collaboration across government, industry, academia, and the volunteer participants necessary to make major advances in human-machine interfaces.

Luke Osborn: Neuromorphic encoding is necessary for reducing complex, high dimensional sensor signals (e.g., tactile, vision, auditory) into meaningful features for driving useful sensory feedback stimulation paradigms in sensory-enabled prostheses, brain-computer interfaces, and human-machine interfaces. A key challenge is understanding what these meaningful features are and how to effectively communicate them back to a human or intelligent system.

Nitish Thakor: Prostheses, whether upper or lower limb for amputees, have concentrated on achieving mechatronic motion and control. Control is achieved through muscle, nerve or direct cortical interface, and using either noninvasive or invasive approaches. However, prosthesis in general and neuroprosthesis in particular are only now attending to incorporating sensory encoding and feedback. I will discuss the need to develop different sensory modalities and mimicking the sensory modalities in biomimetic manner. This would involve sensors that mimic receptors in the skin and encode them as neural spiking activity. This approach has many advantages in terms of encoding, but also poses challenges in terms of novel algorithms and novel approaches to neural interface and feedback. Therefore, my goal is to present some ideas and provoke discussion on sensors, sensory encoding, neural interface and achieving sensory prosthesis solutions for building future closed-loop brain-machine interface technologies.

*Topic 3 - Materials for Neuromorphics (Devices)*
Gert Cauwenberghs: Our work offers an efficient neuromorphic hardware-software co-design framework rectifying the energy-efficiency imbalance between the training and the inference phases in machine learning (ML) artificial intelligence (AI) systems. It leverages the great tolerance of task performance achievable during inference to memory leakage during training, to implement neuromorphic paradigms for energy-efficient memory throughout the learning phase. Hence it substantially reduces the overall energy footprint of ML-based AI, as it is dominated by the energy costs of training which requires significantly more parameter updates and memory writes compared to inference. Operating at multiple time-scales to implement synaptic metaplasticity where fast episodic-memory work in conjunction with one-shot continual online learning architectures, these learning-in-memory systems strike a balance between the adaptation rate of the synaptic elements and their parameter retention capability to realize an optimal traded-off with respect to the energy dissipated per memory write.

Shantanu Chakrabartty: Hybrid neural interfaces, comprising biological and synthetic neurons, provide a pathway for designing the next generation of neuroengineering systems. Existing wetware-hardware implementations have non-overlapping architectural and operational boundaries, with each subsystem optimized independently using platform-specific signal representations. In contrast, a true hybrid neural interface should facilitate cross-mapping between biological and synthetic neurons, enabling the system to seamlessly optimize and delegate computational capabilities between the two domains. This type of interface would not only better control the capabilities of biological systems at the neural circuit level but also augment and repair neural functions at the system level. Currently, several challenges exist in achieving such a hybrid interface. Besides the difficulties in designing biomaterials, electrodes, and optrodes that can seamlessly stimulate and record from biological neurons over long periods, and the challenge of long-term powering of such systems, a critical computational challenge lies in the mismatch at the interface between the neurobiological and neuromorphic domains. First, there is a bandwidth and interface challenge due to the lack of specificity in stimulation and recording, with efficacy that drifts over time. Second, there is a representation challenge arising from the mismatch between the complex spatiotemporal representations in coupled neuronal systems and the simpler spatiotemporal dynamics in current neuromorphic systems. These computational challenges prevent effective cross-mapping between biological and synthetic neurons. We believe that employing an energy-based or Hamiltonian-like joint modeling and synthesis of biological and neuromorphic systems might address these challenges. Such a hybrid neural interface could then continuously evolve and delegate functions across both neurobiology and neuromorphic systems.

Ulkuhan Guler: Biomedical sensors generate continuous streams of data, often requiring real-time processing to ensure timely, accurate analysis and prompt action based on the interpretation of this information. By implementing neuromorphic techniques directly on-chip, data can be processed efficiently with low power consumption, enabling wireless transmission of the processed signals. This reduces the need for off-chip data handling, resulting in compact, energy-efficient, and highly responsive biomedical systems, which are crucial for wearable or implantable health monitoring devices.

Duygu Kuzum: The next leap in implantable neural interfaces requires technological advances in materials, devices, and computing paradigms. Multimodal approaches integrating optical and electrical sensing modalities can overcome spatiotemporal resolution limits of neural sensing as well as open up new avenues for non-invasive neural recording. Integration of sensing, computation and memory on a single array can enable real-time processing of neural signals for compact, low-power and high-throughput neuromorphic brain-machine interfaces. Here, I will present this vision, its challenges, and discuss recent advances in the areas of transparent neural interfaces for multimodal recordings, neuromorphic approaches for on-chip neural processing and computational co-design at the system level for minimally invasive neural interfaces.

Tony Lewis: Intelligent devices, from wearable health monitors to implants, increasingly surround us. As these technologies evolve, more efficient AI systems are crucial. Traditional deep learning, while powerful, struggles with edge deployment due to high computational and energy demands.  We introduce Flow Machines, a novel paradigm for biomedical wearables leveraging state space models and neuromorphic computing. Unlike conventional frame-based deep learning, Flow Machines create internal representations of the external world incrementally, utilizing processing history. This approach contrasts with typical edge-deployed deep learning networks that recompute from scratch, ignoring past computations. Flow Machines accumulate internal representations over time, enabling more compact and efficient AI models suitable for resource-constrained wearables. By transcending traditional paradigms, we aim to enable sophisticated analysis and decision-making in a new class of intelligent biomedical devices.  Our research co-optimizes Flow Machines using state space models as their core architecture. This method addresses edge computing limitations by leveraging internal state and incremental learning, achieving state-of-the-art performance across various edge challenges, from health monitoring to adaptive therapeutics. Flow Machines promise to revolutionize wearable healthcare technology, paving the way for personalized, context-aware solutions and improved patient outcomes.

*Topic 4 - Medical Imaging, Wearables and Analysis*
Brad Aimone: Inspired by experiences in modeling adult hippocampal neurogenesis, in this talk I will describe how the challenges of modeling the diverse time and spatial scales of the brain's plasticity and dynamics led me to look towards the emerging field of neuromorphic computing as a potential solution.

Today's neuromorphic systems are beginning to reach brain-like scales in terms of numbers of neurons, and interacting with these systems presents similar challenges. I will describe a range of unexpected results that we have learned from brain-like computing systems operating at this scale. Finally, I will revisit my original goal of how understanding neural computations at biologically realistic scales on neuromorphic systems can, in turn, help us understand how and why the brain operates the way that it does.

Sihong Wang: Precision medicine, the future of healthcare, promises personalized diagnosis and treatment by considering each individual's unique genetic makeup, age, medical history, and environment. Achieving this vision hinges on two critical capabilities: (1) continuous, high-resolution, multi-modal biosensing at the interface of skin and tissue, and (2) real-time, high-throughput analysis of large, complex datasets to enable timely interventions. However, these advances come with significant challenges. First, bio-electronic interfaces must establish and maintain stable, long-term contact with biological tissues to ensure effective sensing. Second, achieving efficient data processing and power management requires the seamless integration of neuromorphic computing directly onto the bio-electronic sensing platform, enabling edge computing near the sensor. In this talk, I will present our work on developing organic semiconductor- and transistor-based biosensors that incorporate biomimetic properties such as skin-like stretchability, bioadhesiveness, and immune compatibility. I will also introduce our progress in creating intrinsically stretchable neuromorphic devices and arrays capable of performing near-sensor edge computing, advancing the field of bio-electronic interfaces toward real-time, personalized healthcare. The successful deployment of these technologies could revolutionize patient care, enabling earlier detection of diseases and more precise interventions. Realizing this future will require significant investment in cross-disciplinary research, infrastructure, and long-term support for scalable bio-electronic manufacturing.

## Disclaimers

## Acknowledgement

The NPBH Workshop was supported by NSF Conference Award ([2433739](#)) through co-funding from the following NSF programs: Electronics, Photonics & Magnetic Devices (EPMD), Disability and Rehabilitation Engineering (DARE), Communications, Circuits & Sensing-Systems (CCSS), and Foundations of Emerging Technologies (FET). NIH NINDS and NIBIB co-sponsored this workshop via travel support.

The authors acknowledge thoughtful comments from Holly Moore, Vicky Whittemore, Ralph Etienne-Cummings, David Wyatt, Andreas Andreou, Luke Osborn, Tobi Delbruck, Andrea Beckel-Mitchener, John Ngai, Steven Zehnder, Chou Hung, Paul Lane, Soo-Siang Lim, Jeffrey Kopsick, Michele Ferrante, Yvonne

Bennett, Dana Schloesser and reference help from Megan Frankowski. The authors further thank the members of the NIH BRAIN NeuroAI Working Group and the Federal NeuroAI Funders Working Group.

## Generative AI Disclosure

The Anthropic Claude Sonnet 3.7 large-language model was used to generate the initial draft of Appendix 1 based on the corrected audio transcripts of presentations and discussions at the workshop. The draft was subsequently reviewed, edited, and verified for accuracy by the authors by cross-referencing claims, perspectives, and other statements against the transcripts, slides, and other sources. In addition, the authors invited the workshop speakers to review their respective sections in Appendix 1 and to provide feedback to verify accuracy, remedy any omissions, and improve the overall quality of the presentation summaries. The authors thank the workshop participants for contributing their feedback and helping ensure that their views are accurately reflected.